\documentclass{article}

\usepackage[accepted]{style/icml2026}
\usepackage{graphicx}
\usepackage{float}
\usepackage{booktabs}
\usepackage{caption}
\usepackage{hyperref}
\usepackage{amsmath}
\usepackage{amssymb}
\usepackage{mathtools}
\usepackage{amsthm}
\usepackage[capitalize,noabbrev]{cleveref}

\theoremstyle{plain}

\theoremstyle{definition}

\theoremstyle{remark}

\usepackage[textsize=tiny]{todonotes}

\icmltitlerunning{(Sparse) Attention to the Details}

\begin{document}

\twocolumn[
\icmltitle{(Sparse) Attention to the Details: \\ Preserving Spectral Fidelity in ML-based Weather Forecasting Models}

\begin{icmlauthorlist}
\icmlauthor{Maksim Zhdanov}{amlab,uva}
\icmlauthor{Ana Lucic}{uva}
\icmlauthor{Max Welling}{amlab,uva,cusp}
\icmlauthor{Jan-Willem van de Meent}{amlab,uva}
\end{icmlauthorlist}

\icmlaffiliation{amlab}{AMLab}
\icmlaffiliation{uva}{University of Amsterdam}
\icmlaffiliation{cusp}{CuspAI}

\icmlcorrespondingauthor{Maksim Zhdanov}{m.zhdanov@uva.nl}

\icmlkeywords{Machine Learning, Deep Learning, Weather Forecasting, Sparse Attention, Subquadratic Architectures, Long Context, ICML}

\vskip 0.3in
]

\printAffiliationsAndNotice{}

\begin{abstract}
We introduce \textsc{Mosaic}, a probabilistic weather forecasting model that addresses three failure modes of spectral degradation in ML-based weather prediction: spectral damping (statistical), high-frequency aliasing (architectural), and residual high-frequency leakage (parametric). \textsc{Mosaic} generates ensemble members through learned functional perturbations and operates on native-resolution grids via mesh-aligned block-sparse attention, a hardware-aligned mechanism that captures long-range dependencies at linear cost by sharing keys and values across spatially adjacent queries. At 1.5° resolution with 214M parameters, \textsc{Mosaic} matches or outperforms models trained on 6$\times$ finer resolution on key variables and achieves state-of-the-art results among 1.5° models, producing well-calibrated ensembles whose individual members exhibit near-perfect spectral alignment across all resolved frequencies. A 24-member, 10-day forecast takes under 12\,s on a single H100~GPU. Code is available at \href{https://github.com/maxxxzdn/mosaic}{\texttt{github.com/maxxxzdn/mosaic}}.
\end{abstract}
\section{Introduction}
\label{section:introduction}

Accurate weather forecasts save lives and enable timely decisions during extreme events. Phenomena such as frontal zones and tropical cyclones cause catastrophic damage and span relatively short distances, requiring models that faithfully resolve fine spatial scales. Numerical weather prediction (NWP) systems achieve this by integrating the equations of fluid dynamics, thermodynamics, and radiative transfer, but at a computational cost that scales cubically with grid resolution.

\begin{figure}
    \centering
    \input{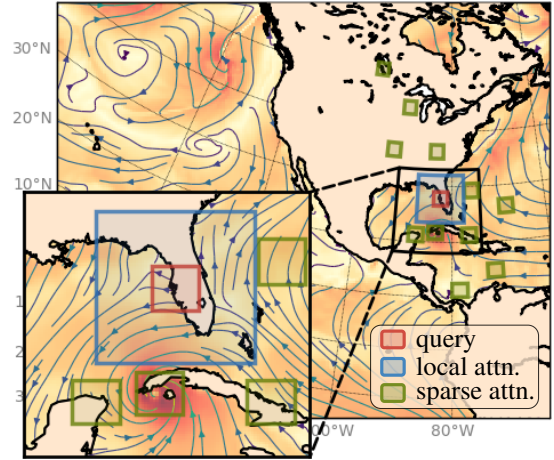}
    \caption{Block-sparse attention for weather forecasting. Spatially close query tokens (red block over Tampa Bay) collectively attend to both local key-value pairs (blue block over Florida) and dynamically selected, spatially distributed ones (green blocks). Sparse attention enables capturing long-range dependencies in high-resolution weather data, critical for extreme events such as hurricane formation (note the eye visible in the inset).}
    \vspace{-0.7em}
    \label{fig:abstract_fig}
\end{figure}

ML-based weather prediction models (MLWPs; \citealt{bi2023accurate, lam2023learning, bodnarFoundation2025}) have emerged as efficient surrogates, generating 10-day forecasts in under 60 seconds on a single GPU, achieving a 1000-10000$\times$ speedup over NWP \cite{buizza1028ifsupgrade, bauer2020ecmwf}. However, these models fail to faithfully reproduce the spectral signature of weather data at fine scales~\cite{mlwp_limit}, which happens for two main reasons.
First, most MLWPs are deterministic and are trained to approximate the conditional mean over future states, which is inherently smoother than any individual realization. Probabilistic MLWPs \cite{gencast, Lang2024AIFSCRPSEF, fgn} address this limitation by producing ensemble members rather than a single mean prediction. Each member represents a plausible realization that can exhibit the sharp, fine-scale structures with spectral properties substantially closer to ground truth than their deterministic counterparts.

\begin{figure*}[t]
    \centering
    \input{figures/spectra_intro}
    \vspace{-0.1em}
    \caption{Spectral analysis and efficiency of MLWP models. \textbf{(a)} Spectral power ratios (model / ERA5) of 10-meter wind speed at $1.5^\circ$ resolution, aggregated over 720 initial conditions throughout the 2020 test year (00:00 and 12:00 UTC) at 24\,h lead time, with 16 ensemble members for probabilistic models. \textbf{(b)} Same quantity at $0.25^\circ$ (HRES-fc0) from a single 6\,h forecast. Probabilistic models (\textsc{Mosaic}, \textsc{ArchesGen}, \textsc{GenCast}) track the reference closely, while deterministic models (\textsc{Stormer}, \textsc{ArchesWeather-Mx4}, \textsc{GraphCast}, \textsc{Aurora}, \textsc{Pangu-Weather}) suppress fine-scale energy. \textsc{Mosaic-C} (a compressed variant entering at a coarser latent grid) and \textsc{Mosaic-R} (a residual variant predicting $x_{t+1}-x_t$ rather than $x_{t+1}$ directly) instead exhibit the opposite signature, with energy rising back toward $1.0$ at the highest wavenumbers, driven by two distinct failure modes: high-frequency aliasing for \textsc{Mosaic-C}, and residual high-frequency leakage for \textsc{Mosaic-R}. \textbf{(c)} Forecast skill vs.\ inference speed on a single H100 GPU; marker size indicates peak GPU memory. nRMSE is computed at 240\,h lead time at $1.5^\circ$ resolution over a set of key variables (see Appendix~\ref{app:evaluation} for details).}
    \vspace{-0.7em}
    \label{fig:spectra_comparison}
\end{figure*}

Second, most MLWPs use compressive encoding in their model architecture: they project high-resolution weather data onto a coarser latent mesh before processing, with a spatial coarsening factor that typically far exceeds any compensating increase in feature channels. This bottleneck has been shown to cause irreversible information loss in both vision~\cite{wang2025scaling} and weather models~\cite{nguyen2023scaling}. Beyond information loss, compressive encoding introduces a second failure mode: when the latent mesh is too coarse to represent fine-scale atmospheric features, nonlinear operations alias their energy into lower-wavenumber modes, which re-emerge as spurious high-frequency content upon decoding~\cite{stability_neural_ops, spectral_neural_ops, aliasfree_vit}. Unlike deterministic smoothing, which suppresses fine-scale spectral energy, this architectural failure mode amplifies it. This signature is visible in \textsc{GenCast} (Fig.~\ref{fig:spectra_comparison}a), which, despite avoiding deterministic smoothing, exhibits a non-monotone bump in spectral energy near the Nyquist limit.

Finally, the choice of output parametrization introduces a third, parametric failure mode. Predicting tendencies $r(x_t)$ and recovering the next state as $x_{t+1}=x_t+r(x_t)$~\citep{lam2023learning, bi2023accurate} leads to high-frequency error accumulation, as any error persists in $x_t$ with high-frequency errors in $r$ added on top, thus compounding over long autoregressive rollouts~\citep{fourcastnet3}. We refer to the failure mode as residual high-frequency leakage. Predicting the next state directly avoids the carry-over.

In this work, we introduce \textsc{Mosaic}: a probabilistic weather forecasting model designed to address all three failure modes of spectral degradation, namely spectral damping (statistical), high-frequency aliasing (architectural), and residual high-frequency leakage (parametric). First, following \cite{fgn}, we use learned functional perturbations to incorporate uncertainty, producing ensemble forecasts whose individual members preserve realistic spectral variability. Second, we propose block-sparse attention: a hardware-aligned formulation of native sparse attention~\cite{nsa} that exploits the intrinsic locality of physical data by sharing keys and values across spatially adjacent queries, computing block-to-block interactions. Each block dynamically selects which regions of the grid to attend to, capturing arbitrarily long-range dependencies at linear cost. To ensure efficient memory access, we process data on the HEALPix mesh~\cite{Gorski1999TheHP}, which offers contiguous storage and therefore allows us to operate on blocks. Third, \textsc{Mosaic} predicts the next weather state directly rather than the residual, avoiding residual high-frequency leakage in long rollouts. Together, these design choices make \textsc{Mosaic} directly applicable to high-resolution grids, with spatial interactions captured at native resolution before any coarsening occurs.

The main contributions of this work are:
\begin{itemize}[leftmargin=25pt, topsep=-1pt, itemsep=-0.2em]
    \item We propose mesh-aligned block-sparse attention, a sparse attention mechanism that captures long-range dependencies at linear cost by jointly selecting key-value blocks for spatially adjacent queries grouped along a physical mesh. This enables weather models to compute spatial interactions at native resolution.
    \item We introduce \textsc{Mosaic}, a probabilistic weather forecasting model that uses block-sparse attention, learned functional perturbations, and direct next-state prediction to address all three failure modes of spectral degradation in MLWPs: spectral damping (statistical), high-frequency aliasing (architectural), and residual high-frequency leakage (parametric).
    \item We demonstrate that \textsc{Mosaic} at 1.5° resolution matches or outperforms models trained on 6$\times$ finer resolution on headline upper-air variables, produces well-calibrated ensembles with near-perfect spectral alignment, and generates a 24-member, 10-day forecast in under 12\,s on a single H100 GPU.
\end{itemize}

\section{Related Work}
\label{section:related_works}

\subsection{Effective Resolution in Weather Forecasting}

\citet{eff_res_ecmwf} define the effective resolution of a weather prediction model as the smallest spatial scale it fully resolves -- the wavelength at which the model's power spectrum starts to deviate from the ground truth. Multiple studies~\cite{mlwp_limit, leveraging_dd} show that MLWPs fail to reproduce realistic spectra and therefore exhibit low effective resolution, struggling to resolve sharp phenomena such as frontal zones and tropical cyclones that span 50-80~km, despite being trained on data at 28~km spatial resolution. \citet{gupta2025mausam} find systematic underestimation of spectral power at mesoscales (10-100~km) across multiple MLWPs. \citet{li2025exploring} demonstrate that Pangu~\cite{bi2023accurate} underestimates kinetic energy at wavelengths below 1000~km and fails to replicate the characteristic ${-\frac{5}{3}}$ spectral slope of physics-based models.

Several strategies address this spectral degradation. \citet{double_pen_fix} modify the training objective to penalize spectral discrepancy directly, improving GraphCast's effective resolution from 1{,}250 to 160~km. Similarly, \citet{fourcastnet3} optimize in the spectral domain, achieving realistic spectra at subseasonal lead times. Hybrid approaches~\cite{Kochkov2023NeuralGC, leveraging_dd} combine ML with physics-based solvers, leveraging the numerical backbone for fine-scale structure. Post-hoc methods~\cite{pde_refiner, no_p_diffusion} condition diffusion models on smooth predictions to recover high-frequency content. Most directly related to our work, \citet{bano2025regional, nordhagen2025high} operate on the original high-resolution mesh with message-passing neural networks and observe significantly better spectral correspondence. We follow the same principle, but replace the fixed graphs with sparse attention, which dynamically determines interactions based on the current weather state.

\subsection{Compression Effect on Expressivity}
The effect of compressive encoding on model expressivity is well studied in computer vision, where patchification, the core tokenization strategy of vision transformers~\cite{dosovitskiy2020image}, reduces computational cost by compressing spatial information. \citet{wang2025scaling} demonstrate irreversible information loss caused by patchification and show that test loss declines consistently as patch size decreases, reaching optimal performance at 1$\times$1 patches. Moreover, reducing patch size is more beneficial than increasing model parameters, indicating that added capacity cannot compensate for compression-induced information loss. The same pattern holds in weather forecasting, where \citet{nguyen2023scaling} show that decreasing patch size consistently improves forecast accuracy.

For atmospheric data, where fine-scale spatial structure is associated with extreme weather events, this information loss is especially consequential. We avoid the compress-first design: rather than projecting onto a coarse mesh before any spatial mixing, we capture spatial interactions at native resolution via block-sparse attention, with coarsening applied only after the first encoder stage.

\subsection{Aliasing in Neural Architectures}
A complementary line of work studies how compressive architectures interact with the frequency content of their inputs. \citet{stability_neural_ops} analyze aliasing in autoregressive neural operators, showing that nonlinear activations generate modes above the Nyquist limit of the representation, which then fold back onto resolved frequencies and accumulate over rollouts. \citet{spectral_neural_ops} formalize this error and propose representations that avoid it. The same phenomenon arises in vision transformers, where \citet{aliasfree_vit} identify patchification as a source of aliasing and study anti-aliasing strategies. These results frame compressive encoding as a limiting factor in frequency representability, which is especially consequential for weather forecasting, where extreme events manifest at fine spatial scales.

\subsection{Ultra-scale Grid Processing}
Avoiding compression shifts the bottleneck to efficiently processing the original ultra-scale data. Standard attention~\cite{vaswani2017attention} scales quadratically with sequence length; FlashAttention~\cite{flash_attn} reduces the memory cost to linear, but still has quadratic complexity. Linear attention~\cite{yang2024gated, yang2023gated} achieves linear cost by replacing the softmax with a kernel-based approximation that maintains a fixed-size state, but sacrifices the input-dependent selectivity that makes standard attention expressive. Native Sparse Attention~(NSA; \citealt{nsa}) resolves this trade-off by computing the dot product over a dynamically selected subset of key-value pairs per query, retaining softmax expressivity at linear cost.

In scientific computing, the need to process large-scale physical simulations has led to complementary approaches. \citet{holzschuh2025pde} handle ultra-scale grids ($256^3$) by partitioning the domain into non-overlapping subdomains processed in parallel. \citet{alkin2024universal, wu2024transolver} use learnable pooling to compress the input into a coarser representation where the bulk of computation occurs. \citet{zhdanov2025erwin, bsa_catalin} avoid pooling and instead use hierarchical trees to impose structure on irregular data, enabling sparse attention with linear cost.

We adapt sparse attention for high-resolution weather grids but further exploit spatial locality: rather than selecting key-value blocks independently per query, we group spatially proximate queries into blocks that jointly select which regions to attend to. Concurrent and independent of our work, \citet{DBLP:conf/iclr/GuLHCZ26} and \citet{longcat_video} develop closely related block-shared sparse attention for video diffusion, and \citet{proxyattn} for long-context language models; these methods target sequence or regular-grid data, where blocks of contiguous indices already correspond to spatial or temporal neighborhoods. We instead target physical data on the sphere, where this property does not hold on the standard latitude-longitude grid; we recover it by operating on the HEALPix mesh with NESTED indexing, so that contiguous-index blocks coincide with spatial neighborhoods on the sphere. This block-level sparsity aligns well with GPU memory access patterns, enabling efficient processing of grids with hundreds of thousands of points without domain decomposition or lossy pooling.

\section{Theoretical Background}
\label{section:theoretical_background}

\subsection{Problem Formulation}
We frame weather forecasting as sampling from the conditional distribution $p\left(X^t \mid X^{t-1},\ldots,X^{t-H}\right)$ over the next atmospheric state $X^t$ given the history of previous states, where each state consists of both surface- and pressure-level atmospheric variables on a latitude-longitude grid. A $T$-step ensemble forecast is generated by sampling autoregressively:
\vspace{-0.3em}
\begin{equation}
\label{eq:problem}
    X^t \sim p\left(X^t \mid X^{t-1},{\ldots},X^{t-H}\right), \quad t = 1, \ldots, T.
    \vspace{-0.3em}
\end{equation}
Drawing multiple trajectories from this process yields an ensemble that quantifies forecast uncertainty.

\begin{figure}
    \centering
    \input{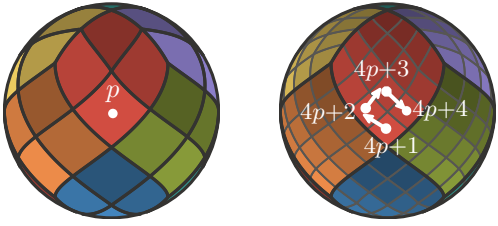}
    \caption{HEALPix mesh refinement. Each pixel (left) is subdivided into four children (right), whose indices follow a Z-order curve that keeps spatially close pixels contiguous in memory.}
    \label{fig:healpix}
    \vspace{-1em}
\end{figure}

\subsection{Hierarchical Sphere Tessellation}
HEALPix~\cite{Gorski1999TheHP} is a hierarchical, equal-area tessellation of a sphere into four-sided polygons (pixels). The sphere is partitioned into a tree-like hierarchy starting from 12 base pixels (4 around each pole, 4 around the equator), with each pixel recursively subdivided into four children, see Fig.~\ref{fig:healpix}. At a given resolution, each pixel covers an identical surface area, ensuring that signal sampling and noise integration are geographically unbiased, unlike traditional latitude-longitude grids, \mbox{which over-sample near the poles.}

\begin{figure*}
    \centering
    \begin{subfigure}[b]{0.24\textwidth}
        \centering
        \caption*{\centering a) \textbf{Interpolation} \par lat-lon $\rightarrow$ HEALPix}
        \includegraphics[width=\textwidth]{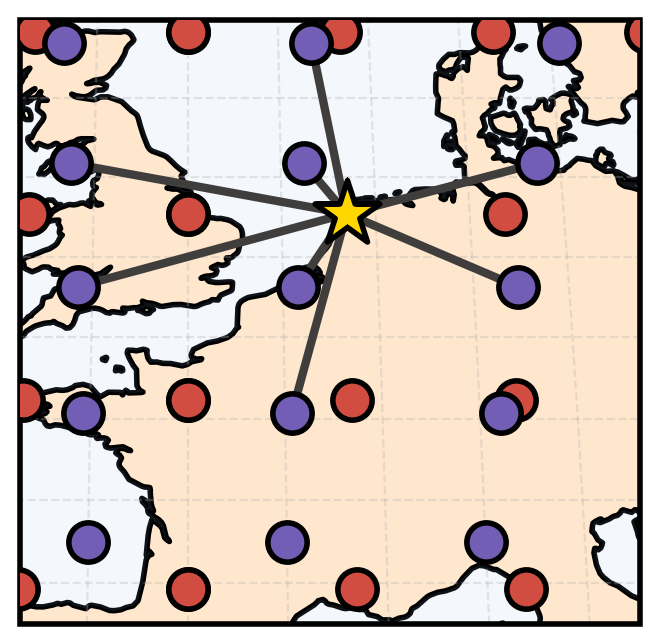}
        \label{fig:sub1}
    \end{subfigure}
    \hfill
    \begin{subfigure}[b]{0.24\textwidth}
        \centering
        \caption*{\centering b) \textbf{Compressed attention} \par between blocks of size 4}
        \includegraphics[width=\textwidth]{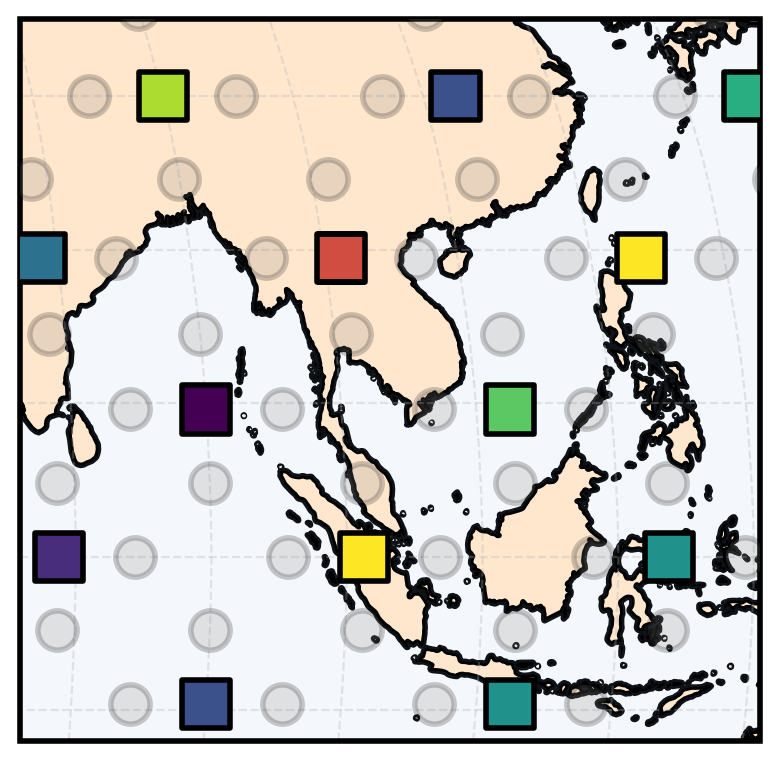}
        \label{fig:sub2}
    \end{subfigure}
    \hfill
    \begin{subfigure}[b]{0.24\textwidth}
        \centering
        \caption*{\centering c) \textbf{Selective attention} \par to top-3 blocks}
        \includegraphics[width=\textwidth]{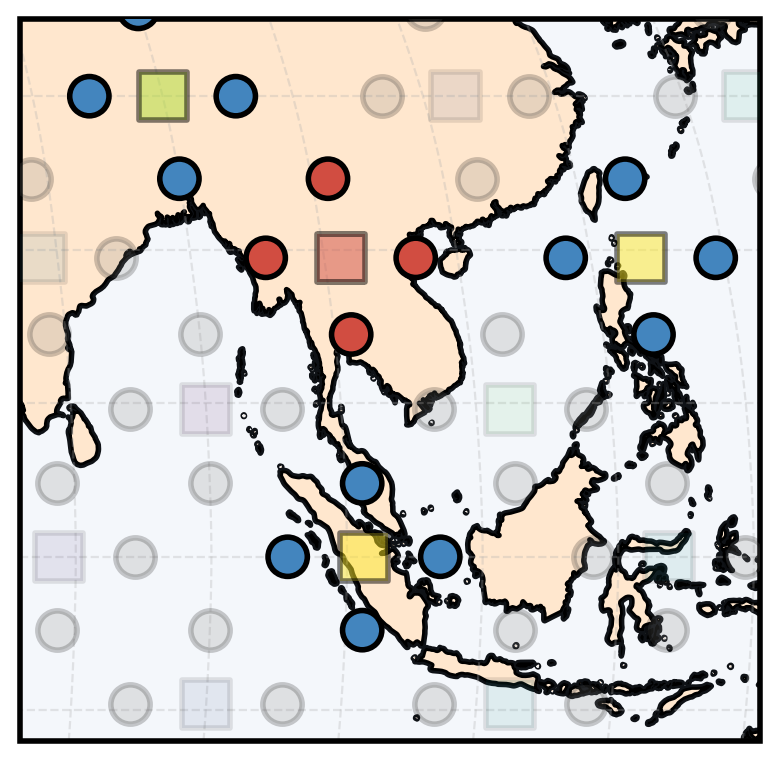}
        \label{fig:sub3}
    \end{subfigure}
    \hfill
    \begin{subfigure}[b]{0.24\textwidth}
        \centering
        \caption*{\centering d) \textbf{Local attention} \par with block size $16$}
        \includegraphics[width=\textwidth]{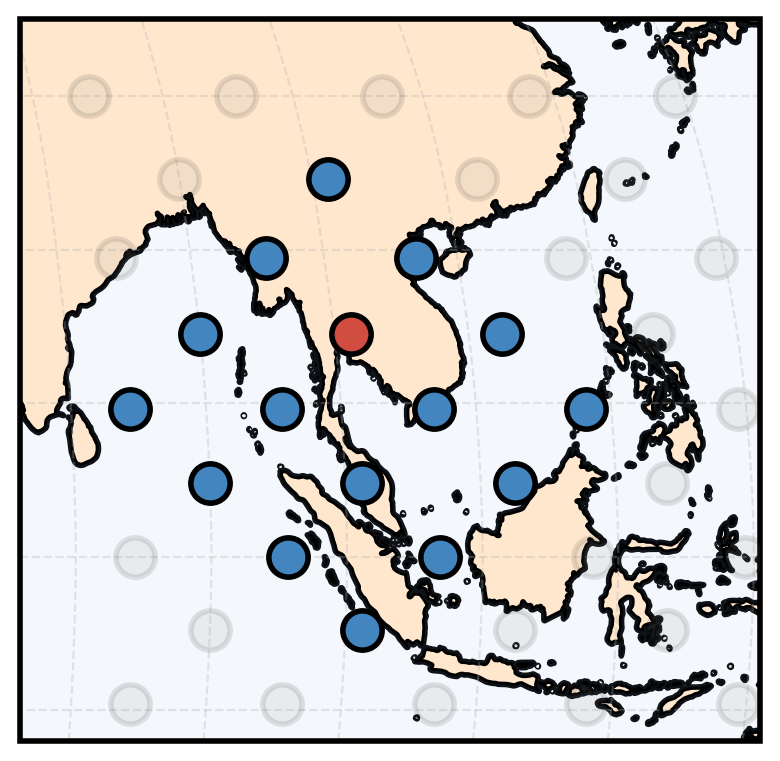}
        \label{fig:sub4}
    \end{subfigure}
    \vspace{-0.5em}
    \caption{Block-sparse attention for weather forecasting. \textbf{(a)} Weather data is interpolated from a latitude-longitude grid (red) to the HEALPix mesh (purple) via cross-attention. \textbf{(b–d)} The three branches of block-sparse attention, illustrated for a single query block (red): \textbf{(b)} Compression computes attention between coarse-grained block representations (squares; color indicates attention score). \textbf{(c)} Selection attends at full resolution to the top-n key blocks. \textbf{(d)} Local attention captures fine-grained interactions within local blocks.}
    \label{fig:main}
\end{figure*}

The size parameter $N_{\textrm{side}}$ defines a HEALPix mesh and determines both the total number of pixels $12 N_{\textrm{side}}^2$ and the approximate angular resolution:
\begin{center}
\begin{tabular}{|l|c|c|c|c|}
\hline
$N_{\textrm{side}}$ & 32 & 64 & 128 & 256 \\ \hline
\# pixels & 12,288 & 49,152 & 196,608 & 786,432 \\ \hline
resolution & $1.83^\circ$ & $0.92^\circ$ & $0.46^\circ$ & $0.23^\circ$ \\ \hline
\end{tabular}
\end{center}

The pixels are organized in memory such that spatially close regions occupy consecutive indices\footnote{This corresponds to the NESTED ordering scheme. We refer the reader to \cite{Gorski1999TheHP} for information about the alternative RING scheme.}. A pixel with index $p$ at resolution $N_{\textrm{side}}$ subdivides into four children with consecutive indices $4p, 4p{+}1, 4p{+}2, 4p{+}3$ at resolution~$2N_{\textrm{side}}$. This contiguous, hierarchical layout makes HEALPix particularly suited for block-based computation: loading a spatially local block of pixels requires a single coalesced memory read rather than the scattered accesses needed on a latitude-longitude grid. Several ML-based weather models already exploit this structure \cite{healvit, pear, dlwp_hpx}. The choice of entry resolution has spectral consequences: halving $N_{\textrm{side}}$ halves the Nyquist wavenumber of the latent grid, so features at finer scales cannot be faithfully represented and their energy may re-emerge as aliased high-frequency content upon decoding (Appendix~\ref{app:aliasing}).

\subsection{Native Sparse Attention}

Native Sparse Attention (NSA; \citet{nsa}) addresses the quadratic complexity of standard self-attention by restricting each query's interactions to a dynamically determined subset of keys and values. The computation is organized into three branches -- compression, selection, and local -- whose outputs $\mathbf{o}^{CG}_i$ (coarse-grained), $\mathbf{o}^{FG}_i$ (fine-grained) and $\mathbf{o}^{L}_i$ (local) are combined via learned gating.
\begin{equation}
\label{eq:aggregated_nsa}
    \mathbf{o}_i = g^{CG}(\mathbf{x}_i) \cdot \mathbf{o}^{CG}_i + g^{FG}(\mathbf{x}_i) \cdot \mathbf{o}^{FG}_i + g^{L}(\mathbf{x}_i) \cdot \mathbf{o}^{L}_i
\end{equation}
where $g^{CG}, g^{FG}, g^{L}$ are learnable linear gating functions.

\paragraph{Compression} Let $\{ B_1, \ldots, B_m \}$ be a partition of tokens into $m$ non-overlapping blocks. For each $B_j$, let $\mathbf{K}_j=\{ \mathbf{k}_l \mid l {\in} B_j \}$ denote the set of keys in the block.~A block representation is computed~via~a~learnable~function~$\varphi$:
\vspace{-0.5em}
\begin{equation}
    \bar{\mathbf{k}}_j = \varphi(\mathbf{K}_j).
\end{equation}
Coarse-grained values are obtained as $\bar{\mathbf{v}}_j = \varphi(\mathbf{V}_j)$ with $\mathbf{V}_j = \{ \mathbf{v}_l \mid l \in B_j \}$. Coarse-grained attention between query $i$ and each block is then evaluated, with attention scores retained for the selection branch:
\begin{equation}
    a_{ij} = \frac{\exp\left( \mathbf{q}_i^\top \bar{\mathbf{k}}_j / \sqrt{d_k} \right)}{\sum_{l=1}^{m} \exp\left( \mathbf{q}_i^\top \bar{\mathbf{k}}_l / \sqrt{d_k} \right)},
    \quad
    \mathbf{o}^{CG}_i = \sum_{j=1}^{m} a_{ij} \bar{\mathbf{v}}_j.
    \vspace{-0.5em}
\end{equation}

The compression branch both captures global context and guides sparsification in the selection branch.

\paragraph{Selection}
For each query $i$, NSA selects the top-$n$ blocks with the highest coarse-grained attention scores, $\mathcal{S}_i = \textrm{top-}n\left(a_{i,:}\right)$, and computes fine-grained attention over keys and values \emph{within} the selected blocks:
\vspace{-0.3em}
\begin{equation}
    \mathbf{o}^{FG}_i = \sum_{j \in \mathcal{S}_i} \sum_{l \in B_j} \frac{\exp\left( \mathbf{q}_i^\top \mathbf{k}_l / \sqrt{d_k} \right)}{Z_i} \mathbf{v}_l
\end{equation}
where $Z_i = \sum_{j \in \mathcal{S}_i} \sum_{l \in B_j} \exp\left( \mathbf{q}_i^\top \mathbf{k}_l / \sqrt{d_k} \right)$ is the normalizing constant. By operating at full resolution, the selection branch preserves fine-scale detail while capturing long-range interactions.

\paragraph{Local Attention}
The local branch applies standard attention for each query $i$ over keys and values within a sliding window, yielding $\mathbf{o}^{L}_i$. This frees the compression and selection branches to focus on long-range interactions.

\section{\textsc{Mosaic}}
\label{section:our_method}

Tobler's first law of geography states that ``nearby things are more related than distant things'' \cite{Tobler1970ACM}. This principle motivates two important design choices in \textsc{Mosaic}: (1) representing data on the HEALPix mesh, where spatially close points occupy contiguous memory, and (2) grouping spatially adjacent tokens into blocks that jointly determine which regions of the globe to attend to.

\subsection{Data Representation on HEALPix}
Weather forecast data is traditionally stored on latitude-longitude grids, where points are ordered row-wise along parallels. This layout places spatially close points far apart in index space, requiring scattered memory accesses to load a block of neighboring pixels. We instead operate on the HEALPix mesh, which guarantees that spatial neighbors occupy contiguous memory locations, enabling coalesced GPU reads and hardware-aligned block computation.

\paragraph{Interpolating Between Grids} To transfer data from the latitude-longitude grid to the HEALPix mesh, we employ the cross-attention interpolation scheme of \cite{DBLP:conf/iclr/WesselsKVPVGB25}. For each target point $i$ on the HEALPix mesh and neighboring source point $j$ on the latitude-longitude grid, queries are computed from the relative position $\mathbf{p}_{ij}$, while keys and values are derived from source features:
\begin{equation}
     \mathbf{q}_{ij} = W_q \, \frac{\mathbf{p}_{ij}}{\|\mathbf{p}_{ij}\|} , \quad\mathbf{k}_j = W_k \mathbf{x}_j, \quad \mathbf{v}_j = W_v \mathbf{x}_j
\end{equation}
The interpolated output at target position $i$ is:
\begin{equation}
\label{eq:interpolation}
     \mathbf{o}_i = \sum_{j \in \mathcal{N}_i} \text{softmax}_j\left( \mathbf{q}_{ij}^\top \mathbf{k}_j / \sqrt{d} \right) \mathbf{v}_j
\end{equation}
\mbox{where $\mathcal{N}_i$ denotes the set of source grid points neighboring~$i$.}

\subsection{Block-Sparse Attention}

BSA retains the three-branch structure of NSA -- compression, selection, and local -- combined via learned gating (Eq.~\ref{eq:aggregated_nsa}). The key difference is that sparsity operates at the block level: rather than each query token independently selecting which key blocks to attend to, tokens within a query block jointly select key blocks. This reduces compression from token-to-block to block-to-block interactions and amortizes the selection cost across all tokens in a block.

\paragraph{Compression} We partition $N$ tokens into $m$ non-overlapping blocks $\{ B_1, \ldots, B_m \}$ and compute coarse-grained representations of queries, keys, and values via a function $\varphi$ (mean pooling in our experiments):
\begin{equation}
     \bar{\mathbf{q}}_i = \varphi(\mathbf{Q}_i), \quad \bar{\mathbf{k}}_j = \varphi(\mathbf{K}_j), \quad \bar{\mathbf{v}}_j = \varphi(\mathbf{V}_j).
\end{equation}
Attention is computed at the block level between query block $i$ and key block $j$:
\begin{equation}
     \bar{a}_{ij} = \frac{\exp\left( \bar{\mathbf{q}}_i^\top \bar{\mathbf{k}}_j / \sqrt{d_k} \right)}{\sum_{l=1}^{m} \exp\left( \bar{\mathbf{q}}_i^\top \bar{\mathbf{k}}_l / \sqrt{d_k} \right)}
\end{equation}
The coarse-grained output is then distributed to all tokens within each query block:
\begin{equation}
     \bar{\mathbf{o}}^{CG}_i = \sum_{j=1}^{m} \bar{a}_{ij} \bar{\mathbf{v}}_j, \qquad \mathbf{o}^{CG}_l = \bar{\mathbf{o}}^{CG}_i \quad \forall l \in B_i.
\end{equation}
Coarse-grained attention captures broad spatial patterns in a computationally feasible manner. For instance, a block over the Netherlands might attend strongly to blocks over the North Atlantic or the Arctic, identifying synoptic-scale influences before the selection branch zooms in on details.

\paragraph{Selection} 
For each query block $i$, the top-$n$ key blocks with the highest coarse-grained attention scores are selected, $\bar{\mathcal{S}}_i = \textrm{top-}n\left(\bar{a}_{i,:}\right)$. Fine-grained attention is then computed between all tokens in query block $i$ and all tokens in the selected key blocks, capturing long-range dependencies:
\begin{equation}
    \mathbf{o}^{FG}_l = \sum_{j \in \bar{\mathcal{S}}_i} \sum_{t \in B_j} \frac{\exp\left( \mathbf{q}_l^\top \mathbf{k}_t / \sqrt{d_k} \right)}{Z_l} \mathbf{v}_t \quad \forall l \in B_i
\end{equation}
where $Z_l = \sum_{j \in \bar{\mathcal{S}}_i} \sum_{t \in B_j} \exp\left( \mathbf{q}_l^\top \mathbf{k}_t / \sqrt{d_k} \right)$ is the normalizing constant. The selection branch resolves fine-scale structure within the chosen regions, for example, by capturing how a specific frontal zone over the Atlantic influences local conditions in Western Europe.

\paragraph{Local Attention} To capture fine-grained local interactions, we compute attention within large blocks of pixels independently. Unlike NSA's sliding window, which would require handling irregular boundaries on the sphere, block attention aligns naturally with the HEALPix structure and can be computed in parallel. The local branch resolves fine-grained structure within each region, such as temperature gradients across a coastline or wind patterns within a valley, freeing other branches to focus on long-range interactions.

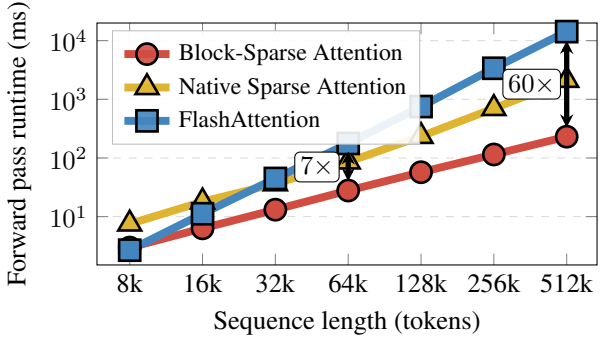
\begin{figure}[t]
    \centering
    \definecolor{myblue}{HTML}{4385BE}
\definecolor{myred}{HTML}{D14D41}
\definecolor{myyellow}{HTML}{DFB431}

\begin{tikzpicture}
\usetikzlibrary{arrows.meta}
\begin{axis}[
    clip=false,
    set layers=standard,
    width=0.48\textwidth,
    height=0.28\textwidth,
    xlabel={Sequence length (tokens)},
    ylabel={Forward pass runtime (ms)},
    xmode=log,
    ymode=log,
    log basis x={2},
    log basis y={10},
    xmin=6000, xmax=700000,
    ymin=1.5, ymax=20000,
    xtick={8192, 16384, 32768, 65536, 131072, 262144, 524288},
    xticklabels={8k, 16k, 32k, 64k, 128k, 256k, 512k},
    ymajorgrids=true,
    ytick={10, 100, 1000, 10000},
    yticklabels={$10^1$, $10^2$, $10^3$, $10^4$},
    minor y tick num=0,
    grid style={gray!30, dashed, line width=0.3pt},
    label style={font=\normalsize},
    tick label style={font=\normalsize},
    legend style={
        at={(0.03, 0.97)},
        anchor=north west,
        font=\small,
        draw=gray!50,
        fill=white,
        fill opacity=0.9,
        text opacity=1,
        inner sep=3pt,
        row sep=1pt,
        cells={anchor=west},
    },
]

\addplot[
    color=myred, line width=3pt, mark=*, mark size=4pt,
    mark options={fill=myred, draw=black, line width=1pt}
] coordinates {
    (8192, 2.97)
    (16384, 6.31)
    (32768, 13.13)
    (65536, 27.84)
    (131072, 57.15)
    (262144, 114.72)
    (524288, 230.68)
};
\addlegendentry{Block-Sparse Attention}

\addplot[
    color=myyellow, line width=3pt, mark=triangle*, mark size=5pt,
    mark options={fill=myyellow, draw=black, line width=1pt}
] coordinates {
    (8192, 7.52)
    (16384, 17.95)
    (32768, 36.49)
    (65536, 86.04)
    (131072, 234.12)
    (262144, 700.32)
    (524288, 2115.22)
};
\addlegendentry{Native Sparse Attention}

\addplot[
    color=myblue, line width=3pt, mark=square*, mark size=4pt,
    mark options={fill=myblue, draw=black, line width=1pt}
] coordinates {
    (8192, 2.68)
    (16384, 11.20)
    (32768, 44.75)
    (65536, 176.85)
    (131072, 752.74)
    (262144, 3383.69)
    (524288, 14258.72)
};
\addlegendentry{FlashAttention}

\begin{pgfonlayer}{axis foreground}
\draw[{Stealth[length=5pt, width=5pt]}-{Stealth[length=5pt, width=5pt]}, black, line width=2pt]
    (axis cs:65536, 40.) -- (axis cs:65536, 130.);
\node[font=\normalsize\bfseries, anchor=east, xshift=-2pt,
    fill=white, draw=black, line width=0.4pt, rounded corners=2pt, inner sep=2pt]
    at (axis cs:63536, {sqrt(27.84*176.85)}) {$7\times$};

\draw[{Stealth[length=5pt, width=5pt]}-{Stealth[length=5pt, width=5pt]}, black, line width=2pt]
    (axis cs:524288, 330.) -- (axis cs:524288, 10000.);
\node[font=\normalsize\bfseries, anchor=east, xshift=-2pt,
    fill=white, draw=black, line width=0.4pt, rounded corners=2pt, inner sep=2pt]
    at (axis cs:522288, {sqrt(230.68*14258.72)}) {$60\times$};
\end{pgfonlayer}

\end{axis}
\end{tikzpicture}
    \caption{Forward pass runtime vs.\ sequence length for Block-Sparse Attention, NSA, and full FlashAttention; measured on NVIDIA RTX A4500. See Appendix~\ref{app:architecture} for details.}
    \label{fig:runtime_scaling}
    \vspace{-1em}
\end{figure}

\paragraph{Computational cost} Let $b$ denote the block size yielding $\frac{N}{b}$ blocks in total. The combined cost of block-sparse attention across branches is:
$
\mathcal{O}\big(N^2/b^2 + Nnb + Nb\big).
$
The compression branch (CG) computes all pairwise interactions between coarse-grained blocks, resulting in $\mathcal{O}(N^2/b^2)$ cost. For block sizes $b \geq 128$, this cost does not form a bottleneck. The selection branch (FG) computes, for each token, attention over $n$ selected blocks of size $b$, yielding $\mathcal{O}(Nnb)$ cost, which is linear in $N$ as $n$ and $b$ are fixed hyperparameters. Local attention is $\mathcal{O}(Nb)$, since each token only attends to tokens within the same block. 
Together, these costs make block-sparse attention scalable to weather grids with hundreds of thousands of tokens. In practice, BSA scales near-linearly with sequence length, achieving up to $61.8\times$ speedup over dense attention and $9.4\times$ over NSA (Fig.~\ref{fig:runtime_scaling}). We implement BSA in Triton~\cite{triton} following the memory-efficient approach of FlashAttention~\cite{flash_attn}; see Appendix~\ref{app:architecture} for details.

\subsection{Model Architecture}
Weather dynamics spans a wide range of spatial scales, from planetary waves spanning thousands of kilometers to mesoscale convective systems tens of kilometers across. To capture this multi-scale structure, we adopt a U-Net architecture that processes data at progressively coarser resolutions in the encoder path, then refines predictions back to the original resolution in the decoder path. Each resolution level consists of block-sparse attention layers that capture interactions at that scale. Skip connections between corresponding encoder and decoder levels preserve fine-grained information that might otherwise be lost during coarsening. Crucially, the first encoder stage computes spatial interactions via block-sparse attention at native resolution before any coarsening occurs. This distinguishes \textsc{Mosaic} from compression-based MLWPs, which first reduce spatial resolution (e.g., via patchification) and then compute interactions on the compressed representation, creating an information bottleneck at the finest scales. By acting on the native grid before coarsening, \textsc{Mosaic} avoids the aliasing that arises when nonlinear operations are applied below the Nyquist limit of the native resolution (Appendix~\ref{app:aliasing}).

\paragraph{Coarsening}
The encoder path progressively coarsens the HEALPix mesh. At each coarsening step, a parent pixel at resolution $N_{\textrm{side}}$ aggregates features from its four children at resolution $2N_{\textrm{side}}$ via learnable pooling:
\begin{equation}
   \mathbf{x}_{\textrm{parent}} = W^{\downarrow}_x \mathbf{X}_c + W^{\downarrow}_p \Delta\mathbf{P}_c,
\end{equation}
where $\mathbf{X}_c = [\mathbf{x}_{c_i}]_{i=1}^{4}$ and $\Delta\mathbf{P}_c = [\Delta\mathbf{p}_{c_i}]_{i=1}^{4}$ are the stacked child features and relative positions, with $\Delta\mathbf{p}_{c_i} = \mathbf{p}_{c_i} - \mathbf{p}_{\textrm{parent}}$. Here $W^{\downarrow}_x, W^{\downarrow}_p$ are learnable projections. At the coarsest resolution, a bottleneck stage applies additional transformer blocks before the decoder path begins.

\paragraph{Refinement}
The decoder path progressively refines features back to the original resolution. At each refinement step, a parent pixel at resolution $N_{\textrm{side}}$ predicts features for its four children at resolution $2N_{\textrm{side}}$:
\begin{equation}
   \mathbf{X}_c = W^{\uparrow}_x \mathbf{x}_{\textrm{parent}} + W^{\uparrow}_p \Delta\mathbf{P}_c.
\end{equation}
The corresponding encoder features are then added to the refined features via skip connections.

\paragraph{Overall architecture}
Input features that include both dynamic (e.g., 2-meter temperature, 10-meter wind components) and static variables (e.g., land-sea mask, soil type) are combined with sinusoidal time (day and year) encodings~\cite{bodnarFoundation2025} and projected to the hidden dimension via a two-layer MLP followed by interpolation to the HEALPix mesh (see Eq.~\ref{eq:interpolation}). 

The data then passes through the U-Net, where a sequence of transformer blocks is applied at each resolution level. Each block has a standard pre-norm structure \cite{llama} with block-sparse attention and SwiGLU \cite{swiglu} feed-forward network:
\begin{equation}
 \begin{split}
    \mathbf{x} &\leftarrow \mathbf{x} + \text{Attention}(\text{RMSNorm}(\mathbf{x})) \\
    \mathbf{x} &\leftarrow \mathbf{x} + \text{FFN}(\text{RMSNorm}(\mathbf{x}), \mathbf{z})
 \end{split}
\end{equation}
We use grouped query attention (GQA; \cite{gqa}) to reduce memory bandwidth requirements, with queries attending to a number of shared key-value heads. We apply 2D rotary positional embeddings (RoPE; \cite{2drope}) to queries and keys, encoding longitude and latitude positions. The head dimension is split equally between the two coordinates, with each half receiving standard RoPE encoding for its respective angle in radians.

After the decoder, features are interpolated back to the latitude-longitude grid and projected to the output dynamic variables, predicting the next weather state $x_{t+1}$ directly rather than the residual tendency $x_{t+1}-x_t$, thereby avoiding residual high-frequency leakage~\citep{fourcastnet3}.

\begin{table*}
  \centering
\begin{small}
\begin{sc}
\begin{tabular}{lccccccccc}
\toprule
                  & \textbf{Res.} & \textbf{Step} & \textbf{s/step}$^\dagger$ & \textbf{Z500} & \textbf{T850} & \textbf{Q700} & \textbf{U850} & \textbf{V850} & \textbf{SP} \\
\midrule
IFS HRES              &    0.1º    & 1h    & ---   & 802.75      & 3.654        & 1848.0       & 6.460        & 6.530        & 748.68      \\
\midrule
Keisler (2022)        & 1º       & 6h     & ---   & 787.01       & 3.559        & 1659.5       & 6.106        & 6.121        & N/A         \\
Stormer ENS (mean)    & 1.4º     & 24h    & 0.06 & 665.88 & 3.001 & 1445.4 & 5.198 & 5.137 & 619.89 \\
ArchesWeather-Mx4     & 1.5º     & 24h    & 0.06 & 693.56       & 3.117        & 1541.1       & 5.413        & 5.431        & 641.08      \\
\midrule
NeuralGCM ENS (50) & 1.4º      & 12h    & 7.42 & \textbf{606.84} & \underline{2.756}   & 1374.2 & \textbf{4.830} & \underline{4.852} & ---  \\
ArchesWeatherGen (mean)& 1.5º  & 24h    & 0.99 & \underline{610.36} & \textbf{2.755} & \underline{1373.7} & \textbf{4.830} & \textbf{4.848} & \textbf{567.15} \\
\midrule
\textsc{Mosaic} (mean; Ours) & 1.5º & 24h & \textbf{0.05} & 624.08 & 2.778 & \textbf{1358.1} & \underline{4.880} & 4.897 & \underline{576.44} \\
\bottomrule
\end{tabular}
\end{sc}
\end{small}
\vspace{1em}
  \caption{Comparison of $1.5^\circ$ resolution models on RMSE scores for key weather variables with 10-day lead-time against ERA5 data, 2020 test year. Best scores in \textbf{bold}, second best \underline{underlined}. Results for additional variables (T2m, U10m, V10m) are provided in Table~\ref{table:main_appendix}. $^\dagger$\,Per-member-per-step inference time (seconds) on a single NVIDIA H100. See Table~\ref{tab:inference_bench} for full benchmarking details.}
  \label{table:main}
\end{table*}

\subsection{Uncertainty Quantification}
Block-sparse attention addresses the architectural source of spectral degradation by capturing spatial interactions at native resolution before any coarsening, avoiding the compress-first bottleneck.
To address the statistical source, \textsc{Mosaic} produces ensemble forecasts via learned functional perturbations \cite{fgn}, a form of Bayesian neural networks \cite{BlundellCKW15}. The key idea is to inject noise into the parameters of the model, resulting in a single weight perturbation affecting the entire forecast in a globally consistent way. \citet{fgn} condition parameters of layer normalization layers on the noise vector; we instead inject noise into SwiGLU layers as bias added inside the gate:
\begin{align}
    \text{cSwiGLU}(x, z) &= \left(\sigma\left(xW_g + z\right) \odot (xW_v)\right) W_{out}
\end{align}
where $\odot$ denotes element-wise multiplication, and $\sigma(x) = \frac{x}{1 + e^{-x}}$ is the Swish activation. We found simple additive noise injection in the SwiGLU gate to work best. Mathematically, this induces an effective output projection $W_{out}^{\text{eff}} = \text{diag}(s) \cdot W_{out}$, where scaling factors $s_i$ are drawn from input-dependent distributions determined by the gate activations $xW_g$. The learned weights $W_{out}$ define the mean structure, while noise $z$ flowing through the input-dependent nonlinearity defines a learned, adaptive covariance. This produces structured uncertainty as the model learns which features should vary across ensemble members based on the input and a small number of latent variables, rather than applying uniform noise everywhere.

\paragraph{Training with CRPS}
The probabilistic model is trained by optimizing the Continuous Ranked Probability Score (CRPS; \citet{fgn}), a proper scoring rule for univariate distributions that encourages both accurate and well-calibrated probabilistic forecasts. We use an unbiased (fair) CRPS estimator \cite{crps}, which for a single variable $y \in \mathbb{R}$ and an ensemble $x^{1:N}$ is given by
\begin{equation}
\label{eq:crps}
\begin{split}
    \text{CRPS}(x^{1:N}, y) &:= \frac{1}{N} \sum_{n} |x^n - y| \\
    &- \frac{1}{2N(N-1)} \sum_{n,n'} |x^n - x^{n'}|.
\end{split}
\end{equation}
The CRPS decomposes into a reliability term penalizing deviation from the ground truth and a sharpness term encouraging ensemble spread only where warranted by true uncertainty. The training loss is the latitude-weighted, variable-weighted fair CRPS averaged over all grid points:
\begin{equation}
\label{eq:loss}
    \mathcal{L} := \frac{1}{|D|} \sum_{d} \frac{1}{HW} \sum_{h,w} \sum_{i=1}^{C} \alpha_i \, \omega_h \, \text{CRPS}(\hat{x}_{i,h,w,d}^{1:N}, \hat{y}_{i,h,w,d})
\end{equation}
where $d$ indexes the batch, $(h,w)$ indexes spatial grid points, $i$ indexes the $C{=}82$ output channels, $\alpha_i$ is a per-channel variable weight~\citep{lam2023learning}, and $\omega_h$ is a latitude weight proportional to $\cos(\phi_h)$. Both predictions $\hat{x}$ and targets $\hat{y}$ are in standardized space. Full details are in Appendix~\ref{app:loss}.

\section{Experiments}
\label{section:experiments}

We aim to answer the following research questions:
\begin{itemize}[leftmargin=20pt, topsep=-1pt, itemsep=-0.2em]
    \item \textbf{RQ1:} How does \textsc{Mosaic} compare against SOTA MLWPs in terms of forecasting skill?
    \item \textbf{RQ2:} Does \textsc{Mosaic} preserve spectral fidelity across all resolved frequencies?
    \item \textbf{RQ3:} Are its ensemble forecasts well-calibrated?
    \item \textbf{RQ4:} Does \textsc{Mosaic}'s spectral fidelity translate into skillful forecasts of real-world extreme events?
\end{itemize}

\begin{figure*}[t]
    \centering
    \usepgfplotslibrary{groupplots}

\tikzset{
    ultra thick/.style={line width=2.0pt}
}

\input{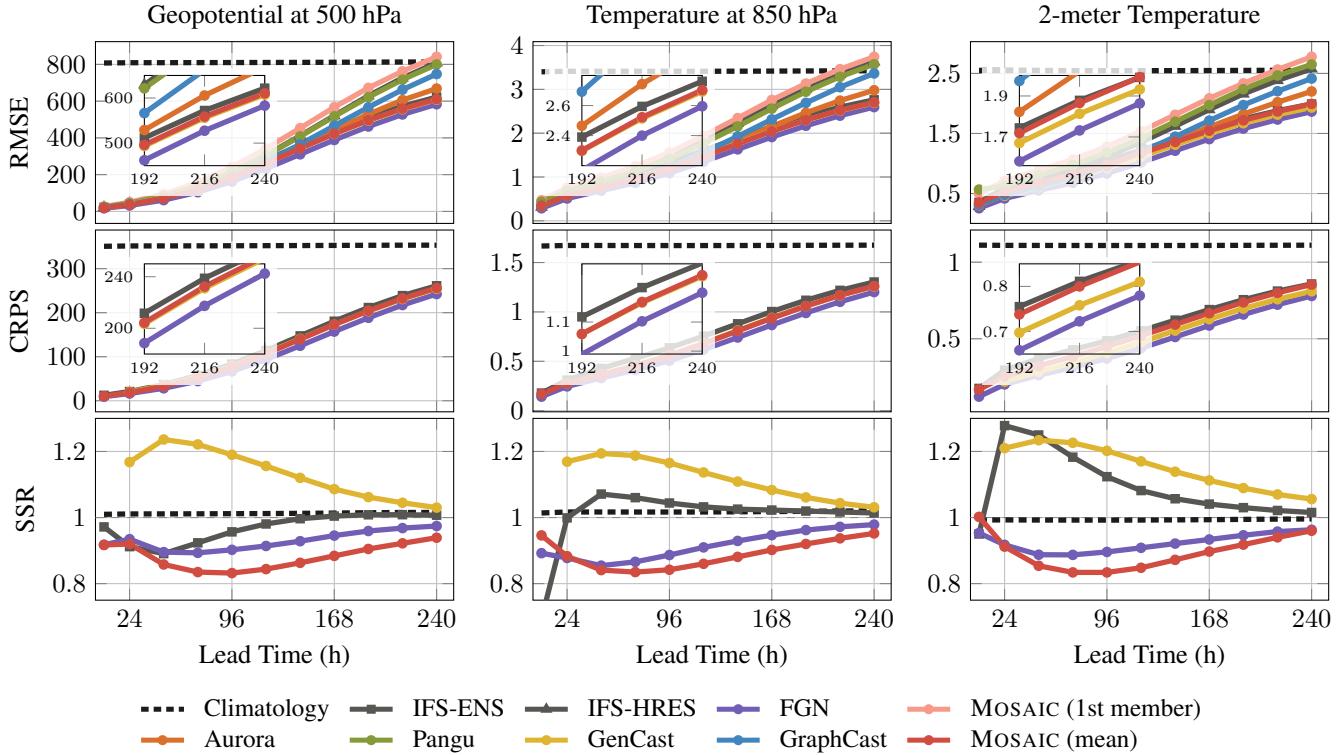}

\definecolor{cBlue}{HTML}{4385BE}
\definecolor{cRed}{HTML}{D14D41}
\definecolor{cGreen}{HTML}{879A39}
\definecolor{cYellow}{HTML}{DFB431}
\definecolor{cOrange}{HTML}{DA702C}
\definecolor{cBlack}{HTML}{1C1B1A}
\definecolor{cGrey}{HTML}{575653}
\definecolor{cPurple}{HTML}{735EB5}
\definecolor{mylightred}{HTML}{F89A8A}

\pgfplotscreateplotcyclelist{modelStyles}{
    {cBlack, densely dashed, ultra thick},
    {cGrey, solid, ultra thick, mark=square*, mark options={fill=white, scale=0.4}},
    {cGrey, solid, ultra thick, mark=triangle*, mark options={fill=white, scale=0.4}},
    {cPurple, solid, ultra thick, mark=*, mark size=1pt},
    {mylightred, solid, ultra thick, mark=*, mark size=1pt}, %
    {cOrange, solid, ultra thick, mark=*, mark size=1pt},
    {cGreen, solid, ultra thick, mark=*, mark size=1pt},
    {cYellow, solid, ultra thick, mark=*, mark size=1pt},
    {cBlue, solid, ultra thick, mark=*, mark size=1pt},
    {cRed, solid, ultra thick, mark=*, mark size=1pt},       %
}

\newcommand{\PlotModelsNoLegend}[1]{
    \addplot table [x=x, y=Clim] {#1};
    \addplot table [x=x, y=IFS_ENS] {#1};
    \addplot table [x=x, y=IFS_HRES] {#1};
    \addplot table [x=x, y=FGN] {#1};
    \addplot table [x=x, y=Ours1] {#1}; %
    \addplot table [x=x, y=Aurora] {#1};
    \addplot table [x=x, y=Pangu] {#1};
    \addplot table [x=x, y=GenCast] {#1};
    \addplot table [x=x, y=GraphCast] {#1};
    \addplot table [x=x, y=Ours] {#1};
}

\newcommand{\PlotModelsInset}[1]{
    \addplot table [x=x, y=IFS_ENS] {#1};
    \addplot table [x=x, y=IFS_HRES] {#1};
    \addplot table [x=x, y=FGN] {#1};
    \addplot table [x=x, y=Ours1] {#1};
    \addplot table [x=x, y=Aurora] {#1};
    \addplot table [x=x, y=Pangu] {#1};
    \addplot table [x=x, y=GenCast] {#1};
    \addplot table [x=x, y=GraphCast] {#1};
    \addplot table [x=x, y=Ours] {#1};
}

\newcommand{\PlotModelsWithLegend}[1]{
    \addplot table [x=x, y=Clim] {#1}; \addlegendentry{Climatology}
    \addplot table [x=x, y=IFS_ENS] {#1}; \addlegendentry{IFS-ENS}
    \addplot table [x=x, y=IFS_HRES] {#1}; \addlegendentry{IFS-HRES}
    \addplot table [x=x, y=FGN] {#1}; \addlegendentry{FGN}
    \addplot table [x=x, y=Ours1] {#1}; \addlegendentry{\textsc{Mosaic} (1st member)}
    \addplot table [x=x, y=Aurora] {#1}; \addlegendentry{Aurora}
    \addplot table [x=x, y=Pangu] {#1}; \addlegendentry{Pangu}
    \addplot table [x=x, y=GenCast] {#1}; \addlegendentry{GenCast}
    \addplot table [x=x, y=GraphCast] {#1}; \addlegendentry{GraphCast}
    \addplot table [x=x, y=Ours] {#1}; \addlegendentry{\textsc{Mosaic} (mean)}
}

\newcommand{\addInset}[3]{
    \node[
        at={(#1.north west)},
        anchor=north west,
        xshift=0.15cm, yshift=-0.35cm,
        minimum width=2.26cm,
        minimum height=1.7cm,
        fill=white,
        fill opacity=0.8,
        inner sep=0pt,
    ] {};

    \begin{axis}[
        scale only axis=true,
        at={(#1.north west)},
        anchor=north west,
        xshift=0.65cm, yshift=-0.45cm,
        width=1.6cm, height=1.2cm,
        xmin=192, xmax=240,
        xtick={192, 216, 240},
        every x tick label/.append style={
            font=\tiny,
            inner sep=1pt,
            yshift=-2pt
        },
        every y tick label/.append style={
            font=\tiny,
            inner sep=4pt,
            yshift=0pt
        },
        cycle list name=modelStyles,
        cycle list shift=1,
        #3
    ]
    \PlotModelsInset{#2}
    \end{axis}
}

\begin{tikzpicture}
    \begin{groupplot}[
        group style={
            group name=plots,
            group size=3 by 3,
            horizontal sep=3em,
            vertical sep=0.25em,
        },
        width=0.37\linewidth,
        height=4cm,
        grid=major,
        xtick={24, 96, 168, 240},
        tick label style={font=\normalsize},
        label style={font=\normalsize},
        title style={font=\normalsize, yshift=-1ex},
        cycle list name=modelStyles,
        unbounded coords=jump,
        xmin=0, xmax=252,
    ]

    \nextgroupplot[
        title={Geopotential at 500 hPa},
        ylabel={RMSE},
        legend columns=5,
        legend to name=sharedlegend,
        legend style={
            font=\normalsize,
            draw=none,
            fill=none,
            column sep=5pt,
            legend cell align={left}
        },
        xticklabels={}
    ]
    \PlotModelsWithLegend{\dataRmseZFive}

    \nextgroupplot[title={Temperature at 850 hPa}, xticklabels={}]
    \PlotModelsNoLegend{\dataRmseTEightFive}

    \nextgroupplot[title={2-meter Temperature}, xticklabels={}, ytick={0.5, 1.5, 2.5},
        yticklabel style={
            /pgf/number format/fixed,
            /pgf/number format/precision=1,
            /pgf/number format/fixed zerofill
        }]
    \PlotModelsNoLegend{\dataRmseTwoMeter}

    \nextgroupplot[ylabel={CRPS}, xticklabels={}]
    \PlotModelsNoLegend{\dataCrpsZFive}

    \nextgroupplot[xticklabels={}]
    \PlotModelsNoLegend{\dataCrpsTEightFive}

    \nextgroupplot[xticklabels={}]
    \PlotModelsNoLegend{\dataCrpsTwoMeter}

    \nextgroupplot[ylabel={SSR}, xlabel={Lead Time (h)}, ytick={0.8, 1.0, 1.2}, ymin=0.75, ymax=1.3]
    \addplot [dashed, color=cBlack!50, thin, forget plot] coordinates {(0,1) (252,1)};
    \PlotModelsNoLegend{\dataSsrZFive}

    \nextgroupplot[xlabel={Lead Time (h)}, ytick={0.8, 1.0, 1.2}, ymin=0.75, ymax=1.3]
    \addplot [dashed, color=cBlack!50, thin, forget plot] coordinates {(0,1) (252,1)};
    \PlotModelsNoLegend{\dataSsrTEightFive}

    \nextgroupplot[xlabel={Lead Time (h)}, ytick={0.8, 1.0, 1.2}, ymin=0.75, ymax=1.3]
    \addplot [dashed, color=cBlack!50, thin, forget plot] coordinates {(0,1) (252,1)};
    \PlotModelsNoLegend{\dataSsrTwoMeter}

    \end{groupplot}

    \addInset{plots c1r1}{\dataRmseZFive}{ymin=450, ymax=650, ytick={500, 600}}
    \addInset{plots c2r1}{\dataRmseTEightFive}{ymin=2.2, ymax=2.8, ytick={2.4, 2.6}}
    \addInset{plots c3r1}{\dataRmseTwoMeter}{ymin=1.56, ymax=2., ytick={1.7, 1.9}}

    \addInset{plots c1r2}{\dataCrpsZFive}{ymin=180, ymax=250, ytick={200, 240}}
    \addInset{plots c2r2}{\dataCrpsTEightFive}{ymin=0.99, ymax=1.3, ytick={1, 1.1}}
    \addInset{plots c3r2}{\dataCrpsTwoMeter}{ymin=0.65, ymax=0.85, ytick={0.7, 0.8}}

    \node[anchor=north] at ($(plots c2r3.south) + (-1.5em,-3em)$) {\scalebox{0.95}{\ref{sharedlegend}}};

\end{tikzpicture}
    \vspace{-1.0em}
    \caption{Forecast skill evaluated against IFS HRES Analysis following the WeatherBench~2 protocol~\citep{wb2} on 2022 test year. Top row: RMSE; middle row: CRPS; bottom row: SSR.}
    \label{fig:rmse}
\end{figure*}

We evaluate \textsc{Mosaic} under two benchmarks that differ in training data, step size, and test year: a controlled 1.5$^\circ$ comparison against other 1.5$^\circ$ models, and a cross-resolution comparison against state-of-the-art 0.25$^\circ$ MLWPs. We further assess extreme-event forecasting through a case study of Hurricane~Ian. We also conduct an ablation study (Appendix~\ref{app:ablation}) where we modify a single aspect of the model relative to a controlled baseline.

\paragraph{Data}
All experiments share the same pretraining data: ERA5 reanalysis from 1979 to 2018 at 1.5$^\circ$ spatial resolution (121$\times$240 equiangular grid). The difference comes down to the finetuning stage:
\begin{itemize}[leftmargin=20pt, topsep=-1pt, itemsep=-0.2em]
    \item \textbf{1.5$^\circ$ benchmark.} Following the protocol of \citet{archesweather}, we finetune on ERA5 2007--2019 to account for distribution shift in the earlier reanalysis period, and test on 2020. Forecasts are evaluated against ERA5 ground truth. \textsc{Mosaic} is trained with a step size of 24\,h to match the baselines.
    \item \textbf{0.25$^\circ$ benchmark.} We finetune on HRES-fc0 analysis from 2016 to 2021 and test on 2022. The evaluation is done by initializing from HRES-fc0, and evaluating against it. \textsc{Mosaic} is trained with a step size of 6\,h as for the majority of the baselines.
\end{itemize}

\paragraph{Baselines}
For the 1.5$^\circ$ benchmark, we compare against models operating at comparable resolution: Stormer~\citep{stormer}, ArchesWeather-Mx4~\citep{archesweather}, ArchesWeatherGen~\citep{archesweather}, NeuralGCM-ENS~\citep{Kochkov2023NeuralGC}. We also include IFS~HRES and IFS~ENS~\citep{ecmwf_medium_range} as NWP reference points. For the 0.25$^\circ$ benchmark, we compare against deterministic models (GraphCast~\citep{lam2023learning}, Aurora~\citep{bodnarFoundation2025}, Pangu-Weather~\citep{bi2023accurate}) and probabilistic models (FGN~\citep{fgn}, GenCast~\citep{gencast}), as well as IFS-ENS, IFS-HRES, and climatology. Training resources of all baselines are given in Table~\ref{tab:model_comparison}.

\begin{figure*}[t]
    \centering
    \begin{subfigure}[t]{0.32\textwidth}
        \centering
        \caption*{\textbf{a)} Init: Sep 23, 2022 12:00 UTC}
        \includegraphics[width=\textwidth]{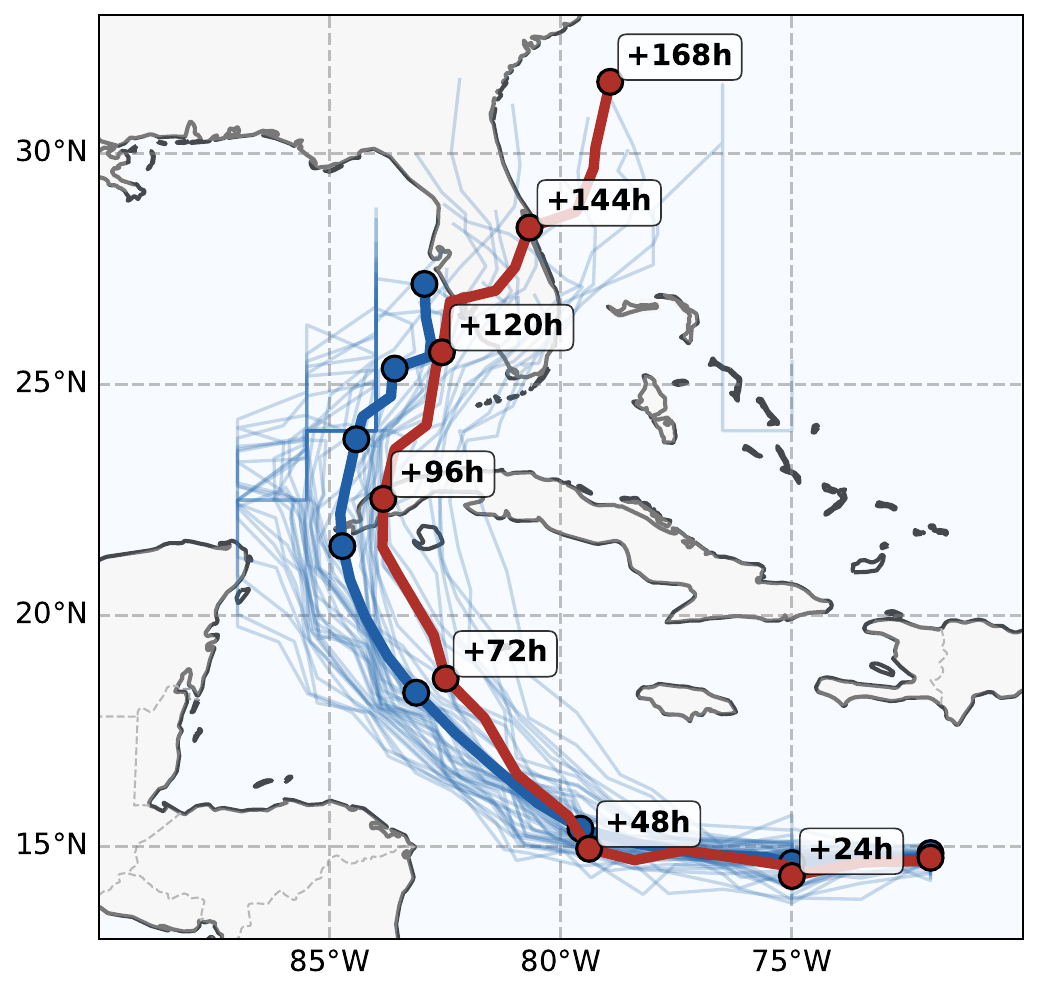}
    \end{subfigure}%
    \hfill
    \begin{subfigure}[t]{0.32\textwidth}
        \centering
        \caption*{\textbf{b)} Init: Sep 25, 2022 12:00 UTC}
        \includegraphics[width=\textwidth]{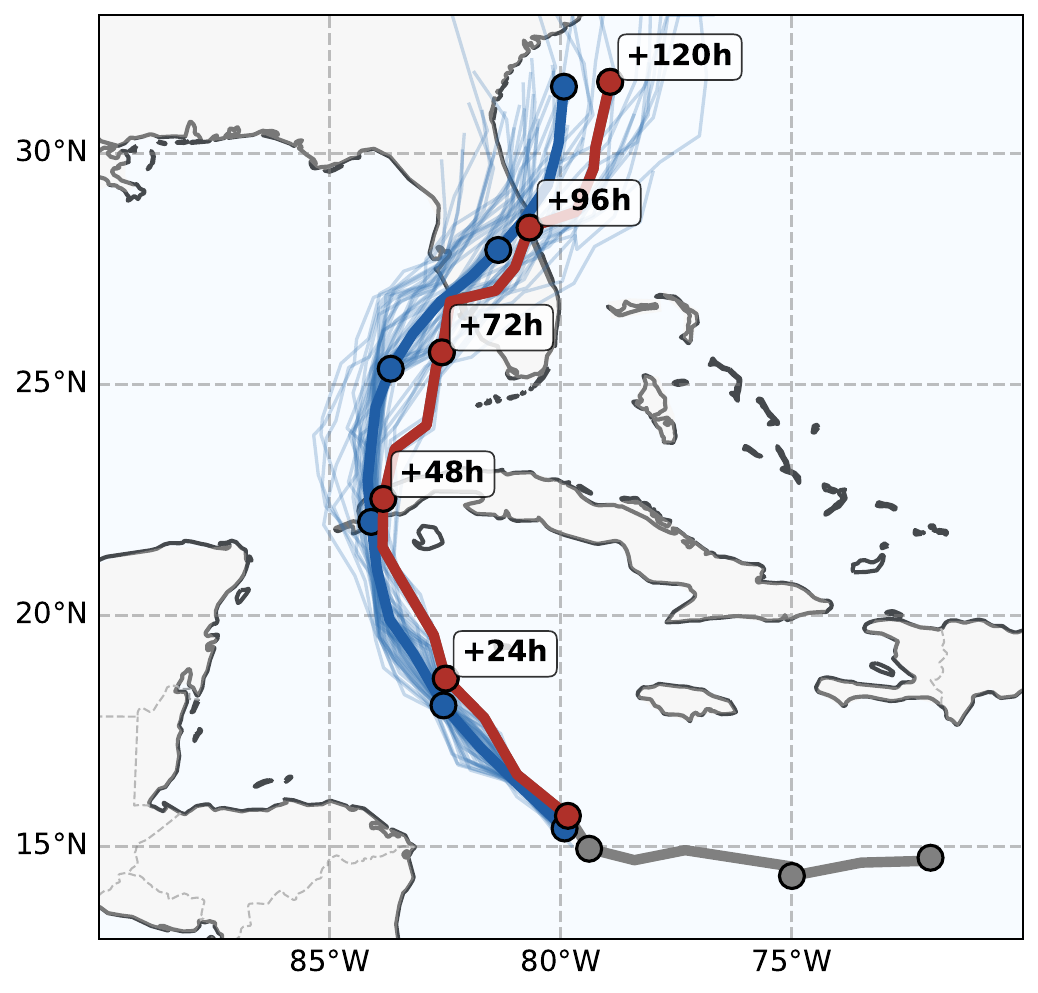}
    \end{subfigure}%
    \hfill
    \begin{subfigure}[t]{0.32\textwidth}
        \centering
        \caption*{\textbf{c)} Init: Sep 27, 2022 12:00 UTC}
        \includegraphics[width=\textwidth]{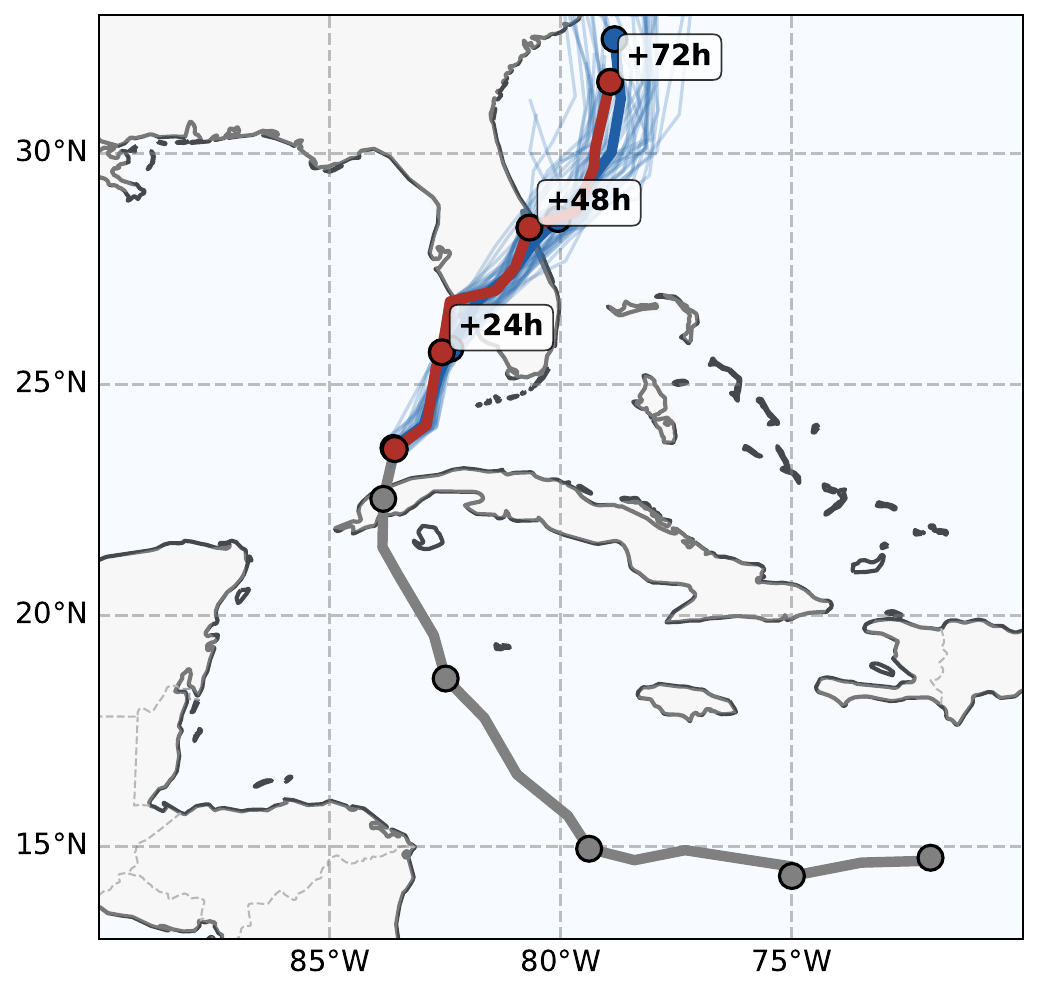}
    \end{subfigure}
    \caption{Hurricane Ian ensemble track forecasts from three initialization times: Sep~23 (7-day lead), Sep~25 (5-day lead), and Sep~27 (3-day lead), all in 2022. Each panel shows 48 ensemble member tracks (light blue), the ensemble mean (dark blue), and the observed track from HRES-fc0 (black). Markers indicate 24\,h intervals. Storm centers are tracked via minimum MSLP guided by IBTrACS best-track positions~\citep{ibtracs}. Wind field evolution is shown in Fig.~\ref{fig:hurricane_ian_evolution}.}
    \label{fig:hurricane_ian_tracks}
\end{figure*}

\paragraph{Training details}
\textsc{Mosaic} contains 214M parameters organized in a U-Net architecture with 14 layers and hidden dimensions progressing from 768 ($N_{\textrm{side}}$=64) to 1024 ($N_{\textrm{side}}$=32) to 1280 ($N_{\textrm{side}}$=16).
Full training is done with a batch size of 2 on 8$\times$ NVIDIA H100 GPUs in fp16 precision. We use the Muon~\citep{muon} optimizer, which is, to our knowledge, the first application of Muon to MLWPs. 
The pretraining stage takes 250,000 steps, while the finetuning phase includes multiple stages with progressively longer autoregressive rollouts (1, 2, 4, 8, 12 steps) for 60k total steps. 
We use the pushforward trick~\citep{brandstetter2022message, bodnarFoundation2025}, where only the final unrolling step is backpropagated through.
During training, we generate ensembles of size 2.

\paragraph{Evaluation}
We follow the WeatherBench~2 protocol~\citep{wb2}: all forecasts are initialized at 00:00 and 12:00 UTC and regridded to 1.5$^\circ$ before computing metrics.
We report latitude-weighted root mean square error (RMSE) of both the first member and the ensemble mean, as well as the fair CRPS~\citep{wb2} of the ensemble (\textbf{RQ1}).
Metrics are computed at each grid point, then globally area-weighted and averaged for each variable and pressure level separately. To evaluate spectral fidelity, we report spectra of global kinetic energy (\textbf{RQ2}).
To evaluate ensemble reliability, we report the spread-to-skill ratio -- the ratio of ensemble spread to the RMSE of the ensemble mean -- where values close to 1 indicate well-calibrated forecasts (\textbf{RQ3}).
To assess whether spectral fidelity translates into skillful extreme event forecasting, we conduct an ensemble case study of Hurricane~Ian using storm-center tracking~\citep{extremeweatherbench} (\textbf{RQ4}).

\subsection*{RQ1: Forecasting skill comparison}

\paragraph{1.5$^\circ$ benchmark}
Table~\ref{table:main} reports 10-day RMSE for all headline variables against ERA5 ground truth.
\textsc{Mosaic} achieves the best Q700 RMSE (1358.1) among all models and the second-best scores on U850 (4.880) and SP (576.44), closely trailing ArchesWeatherGen and NeuralGCM\@.
On Z500, \textsc{Mosaic} (624.08) narrows the gap to the stronger probabilistic baselines NeuralGCM (606.84) and ArchesWeatherGen (610.36), while substantially outperforming the deterministic models Stormer (665.88) and ArchesWeather-Mx4 (693.56).
Notably, \textsc{Mosaic} achieves this performance at a fraction of the cost, being 20$\times$ faster than ArchesWeatherGen and over 150$\times$ faster than NeuralGCM.
As Fig.~\ref{fig:spectra_comparison}\,(c) shows, \textsc{Mosaic} matches the forecast skill of the best probabilistic models while being an order of magnitude faster.

\vspace{-1em}
\paragraph{0.25$^\circ$ benchmark}
\textsc{Mosaic} achieves competitive performance against 0.25$^\circ$ SOTA models.
Fig.~\ref{fig:rmse} shows RMSE and CRPS results across three headline variables: 2-meter temperature (2m temp), geopotential at 500 hPa (Z500), and 10-meter U-component of wind (U10).
At 10-day lead time, \textsc{Mosaic} achieves strong RMSE performance:
on Z500, \textsc{Mosaic} outperforms IFS-ENS (610 vs 622 m$^2$/s$^2$), Aurora (610 vs 668), Pangu (610 vs 800), and GraphCast (610 vs 748);
on U10, \textsc{Mosaic} surpasses IFS-ENS (3.35 vs 3.39 m/s) and all deterministic ML baselines;
on 2m temperature, \textsc{Mosaic} matches IFS-ENS (1.99 K) while outperforming Aurora, Pangu, and GraphCast.

For probabilistic metrics, \textsc{Mosaic} demonstrates competitive CRPS scores.
At 240h, \textsc{Mosaic} achieves Z500 CRPS of 258, close to GenCast (254) and IFS-ENS (261), though FGN achieves the lowest (242).
On U10, \textsc{Mosaic} (1.58) outperforms IFS-ENS (1.61) and approaches GenCast (1.56).
These results demonstrate that block-sparse attention operating at native resolution can extract sufficient information from 1.5$^\circ$ data to compete with models trained on 6$\times$ finer grids.
Tables~\ref{tab:rmse_24h} and~\ref{tab:rmse_240h} provide RMSE at 24\,h and 240\,h for a broader set of headline variables; RMSE, CRPS, and spread-to-skill curves for the remaining variables (V10, MSL, U850, V850, Q700) are shown in Appendix~\ref{appendix:results_025}.
Qualitative unrolling trajectories in Appendix~\ref{app:visualizations} further illustrate \textsc{Mosaic}'s ability to maintain forecast quality over extended lead times.

\subsection*{RQ2: Spectral fidelity preservation}
Fig.~\ref{fig:spectra_comparison}(a) compares kinetic energy spectral ratios of 10-meter wind speed for all ERA5-trained models at 1.5$^\circ$, aggregated over the 2020 test year (720 initial conditions, 16 ensemble members for probabilistic models) at 24\,h lead time. The three failure modes of spectral degradation produce two opposing signatures: spectral damping suppresses fine-scale energy, while high-frequency aliasing and residual high-frequency leakage both amplify it.

Deterministic models exhibit clear suppression at fine scales (Stormer: $0.24$; ArchesWeather-Mx4: $0.47$ at the highest resolved wavenumber), the expected outcome of training against an ensemble mean. \textsc{Mosaic-C}, a compressed variant of \textsc{Mosaic} entering at $N_{\textrm{side}}{=}32$ rather than $64$ (otherwise identical), shows the opposite: its ratio grows past $1.0$ near the Nyquist limit ($1.02$), the signature of aliasing where energy not representable on the coarse latent grid folds back upon decoding. The divergence appears near the native Nyquist rather than the hard limit of the coarser grid because the learnable interpolation absorbs part of the unrepresentable content across the feature dimension, mitigating but not eliminating the bottleneck (Appendix~\ref{app:aliasing}). \textsc{Mosaic-R}, a variant predicting the residual $x_{t+1}-x_t$ rather than $x_{t+1}$ directly (otherwise identical to \textsc{Mosaic}), shows a similar high-wavenumber rise ($1.00$) driven by residual high-frequency leakage rather than aliasing.

\textsc{Mosaic} avoids all three failure modes by combining native-resolution interactions, probabilistic training, and direct next-state prediction, tracking the reference closely ($0.97$). \textsc{ArchesGen} ($0.96$) likewise shows no spectral bump: its deterministic backbone predicts $x_{t+1}$ directly~\citep{archesweather}, with the diffusion component modeling residuals relative to that backbone rather than to $x_t$. Fig.~\ref{fig:spectra_comparison}(b) confirms the same patterns at $0.25^\circ$.

\subsection*{RQ3: Ensemble calibration}
Fig.~\ref{fig:rmse} (bottom row) shows the spread-to-skill ratio across lead times.
Well-calibrated ensembles should exhibit ratios close to 1, with values below 1 indicating overconfidence and values above 1 indicating underconfidence.
\textsc{Mosaic} achieves ratios ranging from 0.83 at medium lead times to 0.95 at 240h, closely matching FGN (0.89--0.97), while GenCast shows persistent underconfidence (1.03--1.24).
This demonstrates that noise injection in SwiGLU gates produces well-calibrated uncertainty estimates on par with state-of-the-art probabilistic MLWPs.

\subsection*{RQ4: Extreme event forecasting: Hurricane Ian}
To evaluate \textsc{Mosaic}'s ability to forecast extreme weather events, we conduct a case study of Hurricane~Ian -- a Category~4 hurricane that made landfall near Fort~Myers, Florida, on September~28, 2022, with sustained winds of 155~mph. We generate 48-member ensembles from three initialization times (Sep~23, 25, and 27, 2022, all at 12Z) and track each member's storm center by locating the minimum mean sea level pressure (MSLP) within a search radius guided by IBTrACS best-track positions~\citep{ibtracs}, following the methodology of ExtremeWeatherBench~\citep{extremeweatherbench}.

Fig.~\ref{fig:hurricane_ian_tracks} shows the resulting ensemble track forecasts, with wind field evolution shown in Fig.~\ref{fig:hurricane_ian_evolution}. At 7-day lead time (Sep~23 init), the ensemble captures the general northward trajectory through the Caribbean, with multiple members already correctly tracking the recurvature over Cuba and landfall on the Florida Gulf Coast. At 5-day lead (Sep~25 init), the ensemble spread narrows, and the ensemble mean closely follows the observed track through recurvature and approach to landfall. At 3-day lead (Sep~27 init), the ensemble tightly brackets the observed track, with most members correctly identifying the landfall region.
Despite operating at 1.5$^\circ$ resolution (${\sim}$166\,km), which cannot resolve the inner-core structure of a tropical cyclone, \textsc{Mosaic} captures the hurricane's evolution. The progressive narrowing of ensemble spread with decreasing lead time demonstrates physically reasonable uncertainty quantification.

\section{Conclusion}
\label{section:conclusion}
We introduced \textsc{Mosaic}, a probabilistic weather forecasting model that addresses three failure modes of spectral degradation in MLWPs: spectral damping from deterministic training, high-frequency aliasing from compressive encoding, and residual high-frequency leakage from residual prediction. \textsc{Mosaic} addresses these through 
(1) learned functional perturbations, which produce ensemble members with realistic spectral variability,
(2) block-sparse attention, which captures spatial interactions at native resolution before any coarsening at linear cost, 
and (3) direct next-state prediction, which avoids carry-over of high-frequency errors in autoregressive rollouts.
\textsc{Mosaic} matches the forecasting skill of state-of-the-art models on both deterministic and probabilistic metrics, while being an order of magnitude faster.
The spectral analysis reveals that individual ensemble members exhibit near-perfect spectral alignment across all resolved frequencies, and a case study of Hurricane~Ian shows that this fidelity translates into skillful ensemble track forecasts of real-world extreme events.

\paragraph{Limitations and Future Work}
\textsc{Mosaic} operates at 1.5$^\circ$ (${\sim}$166\,km), which cannot resolve mesoscale phenomena such as tropical cyclone inner-core structure or individual severe thunderstorms. 
The linear cost and global receptive field of block-sparse attention make scaling to finer grids a natural next step, particularly at 0.25$^\circ$ (${\sim}$700k tokens), where the architectural advantages over compression-based models should be most pronounced. We estimate that the training would require ${\sim}$20{,}000 H100 GPU-hours, which is comparable to existing 0.25$^\circ$ MLWPs. Beyond resolution scaling, \textsc{Mosaic}'s efficiency (12\,s for a 24-member, 10-day forecast on a single GPU) opens the door to real-time ensemble nowcasting, rapid ensemble data assimilation cycles, and operational decision support. Finally, sparse attention over heterogeneous token sets could enable direct ingestion of sparse observations alongside gridded state tokens, potentially bypassing the costly analysis step that current MLWPs require.

\section*{Impact Statement}
This paper presents work whose goal is to advance the field of Machine Learning. There are many potential societal consequences of our work, none which we feel must be specifically highlighted here.

\section*{Acknowledgements}
MZ acknowledges support from Microsoft Research AI4Science.
JWvdM acknowledges support from the European Union Horizon Framework Programme (Grant agreement ID: 101120237).
This work used the Dutch national e-infrastructure with the support of the SURF Cooperative using grant no. EINF-16923.
Computations were partially performed using the University of Amsterdam - Science Faculty (UvA/FNWI) HPC Facility.

\bibliography{bibliography}
\bibliographystyle{style/icml2026}

\newpage
\appendix
\onecolumn
\section{Ablation Study}
\label{app:ablation}

We conduct ablation experiments to validate \textsc{Mosaic}'s key design choices. To make ablations tractable, all variants, including the ablation baseline, are trained under a reduced-scale protocol that differs from the full model (Section~\ref{app:training}) in several ways: training uses ERA5 data from 2007--2018 only (versus 1979--2018), 100k steps with batch size~1 on $4{\times}$ NVIDIA A6000 GPUs (versus 250k steps with batch size~2 on $8{\times}$ H100), single-step rollout only (no multi-step finetuning), and no HRES-fc0 finetuning stage. Each ablation modifies exactly one aspect relative to this reduced baseline.

We report the normalized RMSE (nRMSE): for each variable, the latitude-weighted RMSE of the ensemble-mean prediction is divided by the climatological standard deviation of that variable, yielding a dimensionless error on a comparable scale across all quantities. The reported metric is the mean nRMSE across all 82 output variables, evaluated on 6\,h forecasts over the 2020 test year following the WeatherBench~2 protocol \citep{wb2}.

\begin{table}[htbp]
\centering
\caption{Ablation studies on 6\,h forecast quality (mean nRMSE across all 82 output variables). All variants are trained under a reduced-scale protocol (ERA5 2007--2018, 100k steps, $4{\times}$ A6000, single-step rollout, no HRES-fc0 finetuning). Each ablation modifies exactly one aspect relative to the baseline.}
\label{tab:ablation}
\small
\begin{minipage}[t]{0.32\textwidth}
\raggedright
(a) Model scale. Increasing capacity yields steady improvement.\\[0.5em]
\begin{tabular}{lcc}
\toprule
Variant & Params & nRMSE \\
\midrule
Tiny & 59.8M & 1.0415 \\
Small & 117.9M & 0.1746 \\
Medium & 195.0M & 0.1720 \\
Baseline & 214.3M & 0.1705 \\
\bottomrule
\end{tabular}
\end{minipage}
\hfill
\begin{minipage}[t]{0.32\textwidth}
\raggedright
(b) Sparse attention blocks. More blocks consistently help.\\[0.5em]
\begin{tabular}{lcc}
\toprule
Sparse blocks & nRMSE \\
\midrule
0 / 0 / 0 & 0.1724 \\
12 / 6 / 2 & 0.1708 \\
24 / 12 / 4 & 0.1705 \\
48 / 24 / 8 & 0.1702 \\
\bottomrule
\end{tabular}
\end{minipage}
\hfill
\begin{minipage}[t]{0.32\textwidth}
\raggedright
(c) NSA attention branches. Each branch contributes independently.\\[0.5em]
\begin{tabular}{lc}
\toprule
Branches & nRMSE \\
\midrule
Local only & 0.1724 \\
Local + compression & 0.1720 \\
Local + selection & 0.1723 \\
All three (baseline) & 0.1705 \\
\bottomrule
\end{tabular}
\end{minipage}

\vspace{1.5em}

\begin{minipage}[t]{0.32\textwidth}
\raggedright
(d) Optimizer. Muon outperforms AdamW even with extended budget.\\[0.5em]
\begin{tabular}{lc}
\toprule
Optimizer & nRMSE \\
\midrule
AdamW & 0.1797 \\
AdamW (longer) & 0.1728 \\
Muon (baseline) & 0.1705 \\
\bottomrule
\end{tabular}
\end{minipage}
\hfill
\begin{minipage}[t]{0.32\textwidth}
\raggedright
(e) Training objective. Probabilistic training improves ensemble-mean nRMSE.\\[0.5em]
\begin{tabular}{lc}
\toprule
Objective & nRMSE \\
\midrule
Deterministic (MSE) & 0.1721 \\
Probabilistic (CRPS) & 0.1705 \\
\bottomrule
\end{tabular}
\end{minipage}
\hfill
\begin{minipage}[t]{0.32\textwidth}
\raggedright
(f) History length. Diminishing returns beyond 2 steps.\\[0.5em]
\begin{tabular}{lc}
\toprule
History steps & nRMSE \\
\midrule
1 & 0.1711 \\
2 & 0.1699 \\
4 (baseline) & 0.1705 \\
\bottomrule
\end{tabular}
\end{minipage}

\vspace{1.5em}

\begin{minipage}[t]{0.32\textwidth}
\raggedright
(g) Spatial compression. Avoiding compression preserves forecast quality.\\[0.5em]
\begin{tabular}{lc}
\toprule
Entry $N_{\mathrm{side}}$ & nRMSE \\
\midrule
32 (compressed) & 0.1771 \\
64 (baseline) & 0.1705 \\
\bottomrule
\end{tabular}
\end{minipage}
\hfill
\begin{minipage}[t]{0.32\textwidth}
\raggedright
(h) Rotary position embeddings. RoPE provides a consistent benefit.\\[0.5em]
\begin{tabular}{lc}
\toprule
Variant & nRMSE \\
\midrule
No RoPE & 0.1722 \\
RoPE (baseline) & 0.1705 \\
\bottomrule
\end{tabular}
\end{minipage}
\hfill
\begin{minipage}[t]{0.32\textwidth}
\mbox{}
\end{minipage}
\end{table}

\paragraph{Summary.}
Excluding the undersized Tiny model, all ablation variants fall within a narrow nRMSE band of 0.1699--0.1797, indicating that the architecture is generally robust to individual design changes. We discuss each factor below.

\textit{Model scale\,(a).}
The Tiny configuration (59.8M parameters) diverges to an nRMSE of approximately~1.04, revealing a minimum capacity threshold below which training becomes unstable. Beyond this threshold, scaling from Small (117.9M) to the baseline (214.3M) yields steady gains (0.1746\,$\to$\,0.1705).

\textit{Sparse attention blocks\,(b).}
Increasing the number of sparse blocks consistently improves nRMSE (0.1724\,$\to$\,0.1702), though with diminishing returns. Even without any sparse blocks, performance remains reasonable (0.1724), suggesting that the dense attention layers carry most of the representational load, with sparse blocks providing refinement.

\textit{NSA branches\,(c).}
The three NSA branches -- local, compression, and selection -- contribute complementarily. Local attention alone yields 0.1724, and adding compression alone brings a modest improvement (0.1720), while adding selection alone barely helps (0.1723). The weak contribution of selection in isolation is expected: the selection branch relies on compressed representations to produce its routing scores, so removing the gradients from the compression branch degrades the guidance signal to essentially no useful information. Combining all three branches yields 0.1705, confirming that all branches are necessary.

\textit{Optimizer\,(d).}
Muon outperforms AdamW at equal training budget (0.1705 vs.\ 0.1797). However, the advantage is primarily one of convergence speed: with 50\% more training steps, AdamW closes most of the gap, reaching 0.1728, within 1.3\% of Muon. This makes Muon attractive for the reduced-budget ablation setting, though the two optimizers may converge to more similar performance given sufficient compute.

\textit{Training objective\,(e).}
Probabilistic (CRPS) training improves ensemble-mean nRMSE over deterministic (MSE) training (0.1705 vs.\ 0.1721), consistent with the observation that optimizing for calibrated ensembles also benefits point forecast accuracy.

\textit{History length\,(f).}
Two input time steps yield the best 6\,h nRMSE (0.1699), slightly outperforming the four-step baseline (0.1705). This suggests diminishing returns from longer context at the single-step forecast horizon; the baseline uses four steps to support autoregressive rollouts during finetuning.

\textit{Spatial compression\,(g).}
Entering the U-Net at $N_{\mathrm{side}}$=32 instead of the baseline $N_{\mathrm{side}}$=64 degrades nRMSE by 3.9\% (0.1705\,$\to$\,0.1771), confirming that native-resolution processing in the first encoder stage is important for preserving fine-grained spatial information. The corresponding spectral analysis (Fig.~\ref{fig:spectra_comparison}(a)) shows that this degradation is not a loss of fine-scale detail but the emergence of spurious high-frequency energy near the Nyquist limit -- the signature of aliasing discussed in Appendix~\ref{app:aliasing}.

\textit{Rotary position embeddings\,(h).}
RoPE provides a consistent benefit, reducing nRMSE from 0.1722 to 0.1705, supporting its use for encoding spatial relationships on the HEALPix grid.

Figure~\ref{fig:more_results_combined} extends the main-text evaluation (Fig.~\ref{fig:rmse}) to the remaining surface and pressure-level variables: 10-meter V-wind, mean sea level pressure, U- and V-wind at 850\,hPa, and specific humidity at 700\,hPa. Rows 1--2 show RMSE, rows 3--4 show CRPS, and rows 5--6 show the spread-to-skill ratio (values close to 1.0 indicate well-calibrated ensembles). All forecasts are regridded to 1.5° resolution.

\newpage
\section{Additional Results}
\label{app:visualizations}

\subsection{1.5° Benchmark}
\label{appendix:results_15}

Table~\ref{table:main_appendix} extends the 1.5° comparison from Table~\ref{table:main} to additional headline variables (T850, U850, V850, SP).

\begin{table}[htpb]
  \centering
\begin{small}
\begin{sc}
\begin{tabular}{lccccc}
\toprule
                  & Res. & Step & T2m & U10m & V10m \\
\midrule
IFS HRES              &    0.1º    & 1h      & --          & --          & --          \\
\midrule
Keisler (2022)    & 1º         & 6h       & N/A          & N/A          & N/A          \\
Stormer           & 1.4º       & 24h      & 2.239        & 3.535        & 3.735        \\
NeuralGCM ENS (50) & 1.4º      & 12h      & N/A          & N/A          & N/A          \\
ArchesWeather-Mx4 & 1.5º       & 24h      & 2.323        & 3.716        & 3.924        \\
ArchesWeatherGen (mean) & 1.5º  & 24h      & 2.049 & 3.306 & 3.485 \\
\midrule
\textsc{Mosaic} (mean; Ours) & 1.5º & 24h & 2.053 & 3.372 & 3.559 \\
\bottomrule
\end{tabular}
\end{sc}
\end{small}
\caption{Comparison of coarse-resolution MLWPs on RMSE scores for key weather variables with 240\,h lead time against ERA5 data, 2020 test year (continuation of Table~\ref{table:main}).}
\label{table:main_appendix}
\end{table}

\subsection{0.25° Benchmark}
\label{appendix:results_025}

\begin{figure}[htbp]
    \centering
    \usepgfplotslibrary{groupplots}

\tikzset{
    ultra thick/.style={line width=2.0pt},
    ideal line/.style={dashed, cBlack!50, thin}
}

\input{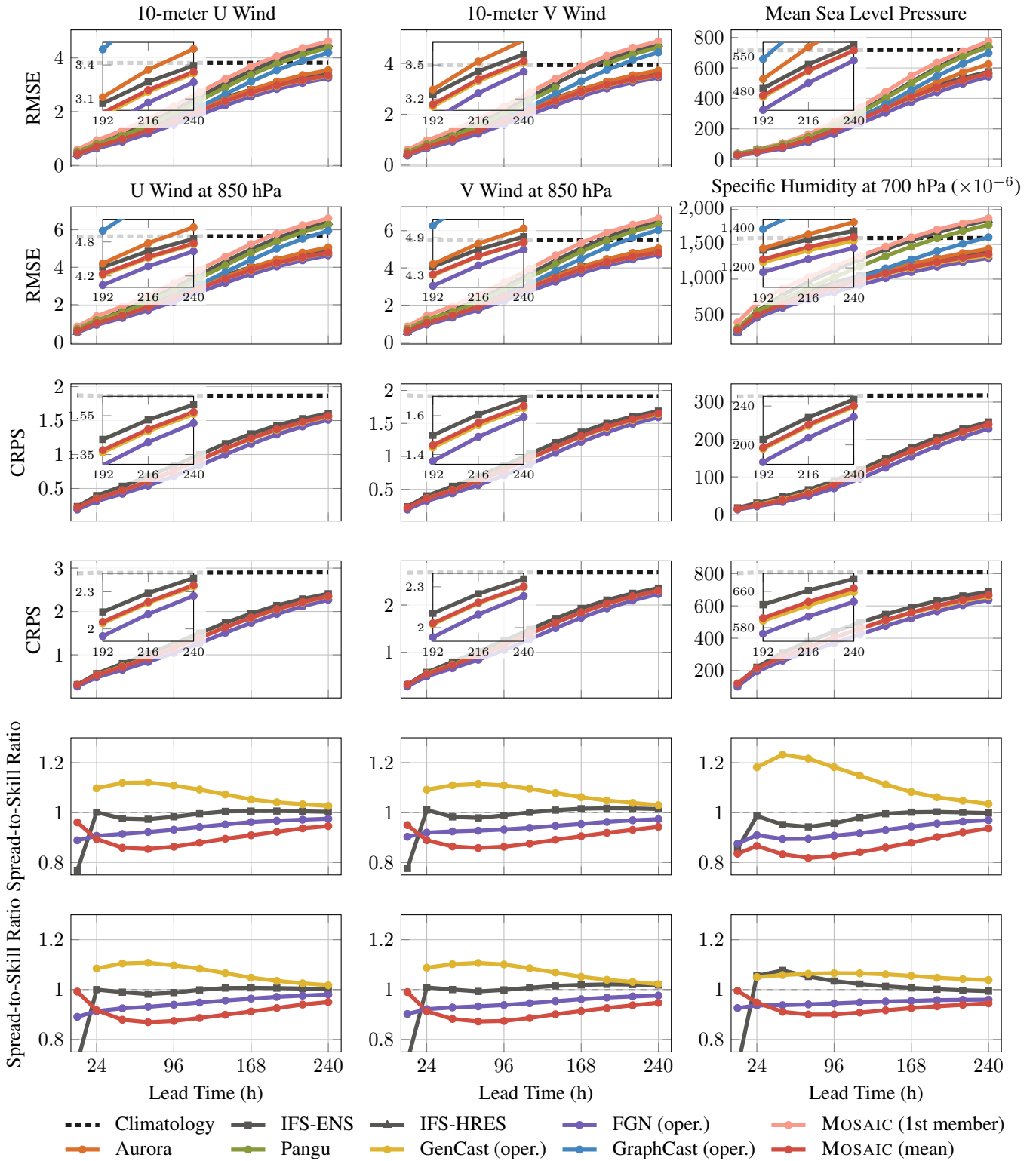}

\definecolor{cBlue}{HTML}{4385BE}
\definecolor{cRed}{HTML}{D14D41}
\definecolor{cGreen}{HTML}{879A39}
\definecolor{cYellow}{HTML}{DFB431}
\definecolor{cOrange}{HTML}{DA702C}
\definecolor{cBlack}{HTML}{1C1B1A}
\definecolor{cGrey}{HTML}{575653}
\definecolor{cPurple}{HTML}{735EB5}
\definecolor{mylightred}{HTML}{F89A8A}

\pgfplotscreateplotcyclelist{modelStyles}{
    {cBlack, densely dashed, ultra thick},
    {cGrey, solid, ultra thick, mark=square*, mark options={fill=white, scale=0.4}},
    {cGrey, solid, ultra thick, mark=triangle*, mark options={fill=white, scale=0.4}},
    {cPurple, solid, ultra thick, mark=*, mark size=1pt},
    {mylightred, solid, ultra thick, mark=*, mark size=1pt}, %
    {cOrange, solid, ultra thick, mark=*, mark size=1pt},
    {cGreen, solid, ultra thick, mark=*, mark size=1pt},
    {cYellow, solid, ultra thick, mark=*, mark size=1pt},
    {cBlue, solid, ultra thick, mark=*, mark size=1pt},
    {cRed, solid, ultra thick, mark=*, mark size=1pt},       %
}

\pgfplotsset{styleIFSENS/.style={cGrey, solid, ultra thick, mark=square*, mark options={fill=white, scale=0.4}}}
\pgfplotsset{styleFGN/.style={cPurple, solid, ultra thick, mark=*, mark size=1pt}}
\pgfplotsset{styleGenCast/.style={cYellow, solid, ultra thick, mark=*, mark size=1pt}}
\pgfplotsset{styleOurs/.style={cRed, solid, ultra thick, mark=*, mark size=1pt}}

\newcommand{\PlotModelsNoLegendCombined}[1]{
    \addplot table [x=x, y=Clim] {#1};
    \addplot table [x=x, y=IFS_ENS] {#1};
    \addplot table [x=x, y=IFS_HRES] {#1};
    \addplot table [x=x, y=FGN] {#1};
    \addplot table [x=x, y=Ours1] {#1};
    \addplot table [x=x, y=Aurora] {#1};
    \addplot table [x=x, y=Pangu] {#1};
    \addplot table [x=x, y=GenCast] {#1};
    \addplot table [x=x, y=GraphCast] {#1};
    \addplot table [x=x, y=Ours] {#1};
}

\newcommand{\PlotModelsWithLegendCombined}[1]{
    \addplot table [x=x, y=Clim] {#1}; \addlegendentry{Climatology}
    \addplot table [x=x, y=IFS_ENS] {#1}; \addlegendentry{IFS-ENS}
    \addplot table [x=x, y=IFS_HRES] {#1}; \addlegendentry{IFS-HRES}
    \addplot table [x=x, y=FGN] {#1}; \addlegendentry{FGN (oper.)}
    \addplot table [x=x, y=Ours1] {#1}; \addlegendentry{\textsc{Mosaic} (1st member)}
    \addplot table [x=x, y=Aurora] {#1}; \addlegendentry{Aurora}
    \addplot table [x=x, y=Pangu] {#1}; \addlegendentry{Pangu}
    \addplot table [x=x, y=GenCast] {#1}; \addlegendentry{GenCast (oper.)}
    \addplot table [x=x, y=GraphCast] {#1}; \addlegendentry{GraphCast (oper.)}
    \addplot table [x=x, y=Ours] {#1}; \addlegendentry{\textsc{Mosaic} (mean)}
}

\newcommand{\PlotModelsInsetCombined}[1]{
    \addplot table [x=x, y=IFS_ENS] {#1};
    \addplot table [x=x, y=IFS_HRES] {#1};
    \addplot table [x=x, y=FGN] {#1};
    \addplot table [x=x, y=Ours1] {#1};
    \addplot table [x=x, y=Aurora] {#1};
    \addplot table [x=x, y=Pangu] {#1};
    \addplot table [x=x, y=GenCast] {#1};
    \addplot table [x=x, y=GraphCast] {#1};
    \addplot table [x=x, y=Ours] {#1};
}

\newcommand{\PlotSSRCombined}[1]{
    \addplot[ideal line] coordinates {(0,1) (252,1)};
    \addplot[styleIFSENS] table [x=x, y=IFS_ENS] {#1};
    \addplot[styleFGN]    table [x=x, y=FGN]     {#1};
    \addplot[styleGenCast] table [x=x, y=GenCast] {#1};
    \addplot[styleOurs]   table [x=x, y=Ours]    {#1};
}

\newcommand{\addInsetCombined}[3]{
    \node[
        at={(#1.south east)},
        anchor=south east,
        xshift=-2.38cm, yshift=0.68cm,
        minimum width=2.26cm,
        minimum height=1.7cm,
        fill=white,
        fill opacity=0.8,
        inner sep=0pt,
    ] {};

    \begin{axis}[
        scale only axis=true,
        at={(#1.south east)},
        anchor=south east,
        xshift=-2.6cm, yshift=1.cm,
        width=1.6cm, height=1.2cm,
        xmin=192, xmax=240,
        xtick={192, 216, 240},
        every x tick label/.append style={
            font=\tiny,
            inner sep=1pt,
            yshift=-2pt
        },
        every y tick label/.append style={
            font=\tiny,
            inner sep=4pt,
            yshift=0pt
        },
        cycle list name=modelStyles,
        cycle list shift=1,
        #3
    ]
    \PlotModelsInsetCombined{#2}
    \end{axis}
}

\begin{tikzpicture}
    \begin{groupplot}[
        group style={
            group name=allplots,
            group size=3 by 6,
            horizontal sep=3em,
            vertical sep=2em,
        },
        width=0.37\linewidth,
        height=4cm,
        grid=major,
        xtick={24, 96, 168, 240},
        tick label style={font=\normalsize},
        label style={font=\normalsize},
        title style={font=\normalsize, yshift=-1ex},
        cycle list name=modelStyles,
        unbounded coords=jump,
        xmin=0, xmax=252,
    ]

    \nextgroupplot[
        title={10-meter U Wind},
        ylabel={RMSE},
        legend columns=5,
        legend to name=combinedlegend,
        legend style={
            font=\normalsize,
            draw=none,
            fill=none,
            column sep=5pt,
            legend cell align={left}
        },
        xticklabels={}
    ]
    \PlotModelsWithLegendCombined{\dataRmseUTenm}

    \nextgroupplot[title={10-meter V Wind}, xticklabels={}]
    \PlotModelsNoLegendCombined{\dataRmseVTenm}

    \nextgroupplot[title={Mean Sea Level Pressure}, xticklabels={}]
    \PlotModelsNoLegendCombined{\dataRmseMsl}

    \nextgroupplot[title={U Wind at 850 hPa}, ylabel={RMSE}, xticklabels={}]
    \PlotModelsNoLegendCombined{\dataRmseUEightFive}

    \nextgroupplot[title={V Wind at 850 hPa}, xticklabels={}]
    \PlotModelsNoLegendCombined{\dataRmseVEightFive}

    \nextgroupplot[title={Specific Humidity at 700 hPa ($\times 10^{-6}$)}, xticklabels={}]
    \PlotModelsNoLegendCombined{\dataRmseQSeven}

    \nextgroupplot[ylabel={CRPS}, xticklabels={}]
    \PlotModelsNoLegendCombined{\dataCrpsUTenm}

    \nextgroupplot[xticklabels={}]
    \PlotModelsNoLegendCombined{\dataCrpsVTenm}

    \nextgroupplot[xticklabels={}]
    \PlotModelsNoLegendCombined{\dataCrpsMsl}

    \nextgroupplot[ylabel={CRPS}, xticklabels={}]
    \PlotModelsNoLegendCombined{\dataCrpsUEightFive}

    \nextgroupplot[xticklabels={}]
    \PlotModelsNoLegendCombined{\dataCrpsVEightFive}

    \nextgroupplot[xticklabels={}]
    \PlotModelsNoLegendCombined{\dataCrpsQSeven}

    \nextgroupplot[
        ylabel={Spread-to-Skill Ratio},
        ytick={0.8, 1.0, 1.2},
        ymin=0.75, ymax=1.3,
        cycle list name=modelStyles,
        xticklabels={}
    ]
    \PlotSSRCombined{\dataSsrUTenm}

    \nextgroupplot[
        ytick={0.8, 1.0, 1.2},
        ymin=0.75, ymax=1.3,
        xticklabels={}
    ]
    \PlotSSRCombined{\dataSsrVTenm}

    \nextgroupplot[
        ytick={0.8, 1.0, 1.2},
        ymin=0.75, ymax=1.3,
        xticklabels={}
    ]
    \PlotSSRCombined{\dataSsrMsl}

    \nextgroupplot[
        ylabel={Spread-to-Skill Ratio},
        xlabel={Lead Time (h)},
        ytick={0.8, 1.0, 1.2},
        ymin=0.75, ymax=1.3,
    ]
    \PlotSSRCombined{\dataSsrUEightFive}

    \nextgroupplot[
        xlabel={Lead Time (h)},
        ytick={0.8, 1.0, 1.2},
        ymin=0.75, ymax=1.3,
    ]
    \PlotSSRCombined{\dataSsrVEightFive}

    \nextgroupplot[
        xlabel={Lead Time (h)},
        ytick={0.8, 1.0, 1.2},
        ymin=0.75, ymax=1.3,
    ]
    \PlotSSRCombined{\dataSsrQSeven}

    \end{groupplot}

    \addInsetCombined{allplots c1r1}{\dataRmseUTenm}{ymin=3.0, ymax=3.6, ytick={3.1, 3.4}}
    \addInsetCombined{allplots c2r1}{\dataRmseVTenm}{ymin=3.1, ymax=3.7, ytick={3.2, 3.5}}
    \addInsetCombined{allplots c3r1}{\dataRmseMsl}{ymin=440, ymax=580, ytick={480, 550}}

    \addInsetCombined{allplots c1r2}{\dataRmseUEightFive}{ymin=4.0, ymax=5.2, ytick={4.2, 4.8}}
    \addInsetCombined{allplots c2r2}{\dataRmseVEightFive}{ymin=4.1, ymax=5.2, ytick={4.3, 4.9}}
    \addInsetCombined{allplots c3r2}{\dataRmseQSeven}{ymin=1100, ymax=1450, ytick={1200, 1400}}

    \addInsetCombined{allplots c1r3}{\dataCrpsUTenm}{ymin=1.3, ymax=1.65, ytick={1.35, 1.55}}
    \addInsetCombined{allplots c2r3}{\dataCrpsVTenm}{ymin=1.35, ymax=1.7, ytick={1.4, 1.6}}
    \addInsetCombined{allplots c3r3}{\dataCrpsMsl}{ymin=180, ymax=250, ytick={200, 240}}

    \addInsetCombined{allplots c1r4}{\dataCrpsUEightFive}{ymin=1.9, ymax=2.45, ytick={2.0, 2.3}}
    \addInsetCombined{allplots c2r4}{\dataCrpsVEightFive}{ymin=1.9, ymax=2.4, ytick={2.0, 2.3}}
    \addInsetCombined{allplots c3r4}{\dataCrpsQSeven}{ymin=550, ymax=700, ytick={580, 660}}

    \node[anchor=north] at ($(allplots c2r6.south) + (0,-2.5em)$) {\ref{combinedlegend}};

\end{tikzpicture}
    \caption{RMSE (rows 1--2), CRPS (rows 3--4), and spread-to-skill ratio (rows 5--6) as a function of lead time for additional variables not shown in the main text. Within each metric, the top row shows surface variables (10-meter U-wind, 10-meter V-wind, mean sea level pressure) and the bottom row shows pressure-level variables (U850, V850, Q700). Values close to 1.0 (dashed line) in the SSR rows indicate well-calibrated ensembles.}
    \label{fig:more_results_combined}
\end{figure}

Tables~\ref{tab:rmse_24h} and~\ref{tab:rmse_240h} report per-variable RMSE at 24\,h and 240\,h lead times for all models evaluated at 1.5° resolution.
Figure~\ref{fig:more_results_combined} extends the 0.25° evaluation from Fig.~\ref{fig:rmse} to the remaining surface and pressure-level variables, showing RMSE, CRPS, and spread-to-skill ratio as a function of lead time.

\begin{table}[htbp]
  \centering
\begin{small}
\begin{sc}
\begin{tabular}{llccccccccc}
\toprule
                  & Res. & Z500 & T850 & Q700 & U850 & V850 & T2m & SP & U10m & V10m \\
\midrule
IFS HRES              & 0.1°  & 41.23 & 0.634 & 537.3 & 1.126 & 1.149 & 0.541 & 58.70 & 0.802 & 0.832 \\
IFS ENS               & 0.2°  & 40.91 & 0.630 & 503.1 & 1.102 & 1.121 & 0.589 & 59.44 & 0.784 & 0.806 \\
\midrule
Pangu                 & 0.25° & 45.01 & 0.681 & 537.2 & 1.180 & 1.218 & 0.624 & 59.54 & 0.796 & 0.825 \\
GraphCast             & 0.25° & 38.32 & 0.570 & 475.3 & 1.021 & 1.048 & 0.476 & 49.36 & 0.692 & 0.719 \\
GenCast               & 0.25° & 39.20 & 0.570 & 480.7 & 1.040 & 1.070 & 0.462 & 50.04 & 0.701 & 0.732 \\
FGN                   & 0.25° & 32.27 & 0.510 & 443.2 & 0.928 & 0.953 & 0.418 & 42.22 & 0.627 & 0.655 \\
Aurora                & 0.25° & 38.12 & 0.566 & 477.9 & 1.003 & 1.031 & 0.474 & 48.10 & 0.681 & 0.710 \\
\midrule
\textsc{Mosaic} (Ours) & 1.5° & 36.78 & 0.572 & 482.8 & 1.028 & 1.054 & 0.525 & 47.91 & 0.701 & 0.729 \\
\bottomrule
\end{tabular}
\end{sc}
\end{small}
\caption{RMSE scores for key weather variables at 24\,h lead time. All metrics are evaluated at 1.5° resolution.}
\label{tab:rmse_24h}
\end{table}

\begin{table}[htbp]
  \centering
\begin{small}
\begin{sc}
\begin{tabular}{llccccccccc}
\toprule
                  & Res. & Z500 & T850 & Q700 & U850 & V850 & T2m & SP & U10m & V10m \\
\midrule
IFS HRES              & 0.1°  & 809.0  & 3.626 & 1847.  & 6.417 & 6.500 & 2.576 & 753.3 & 4.513 & 4.767 \\
IFS ENS               & 0.2°  & 621.9  & 2.761 & 1390.  & 4.854 & 4.918 & 1.991 & 574.7 & 3.394 & 3.595 \\
\midrule
Pangu                 & 0.25° & 799.8  & 3.573 & 1782.  & 6.267 & 6.334 & 2.648 & 742.4 & 4.378 & 4.633 \\
GraphCast             & 0.25° & 747.6  & 3.364 & 1601.  & 5.953 & 6.032 & 2.415 & 698.9 & 4.195 & 4.436 \\
GenCast               & 0.25° & 607.5  & 2.693 & 1340.  & 4.758 & 4.826 & 1.932 & 563.2 & 3.325 & 3.521 \\
FGN                   & 0.25° & 583.1  & 2.595 & 1302.  & 4.636 & 4.710 & 1.864 & 542.7 & 3.247 & 3.440 \\
Aurora                & 0.25° & 668.4  & 2.979 & 1435.  & 5.060 & 5.051 & 2.200 & 624.9 & 3.542 & 3.718 \\
\midrule
\textsc{Mosaic} (Ours) & 1.5° & 610.4  & 2.703 & 1359.  & 4.773 & 4.828 & 1.992 & 563.2 & 3.339 & 3.534 \\
\bottomrule
\end{tabular}
\end{sc}
\end{small}
\caption{RMSE scores for key weather variables at 240\,h (10-day) lead time. All metrics are evaluated at 1.5° resolution.}
\label{tab:rmse_240h}
\end{table}

\subsection{Hurricane Ian Case Study}
\label{app:hurricane_ian}

\begin{figure}[htbp]
    \centering
    \includegraphics[width=\textwidth]{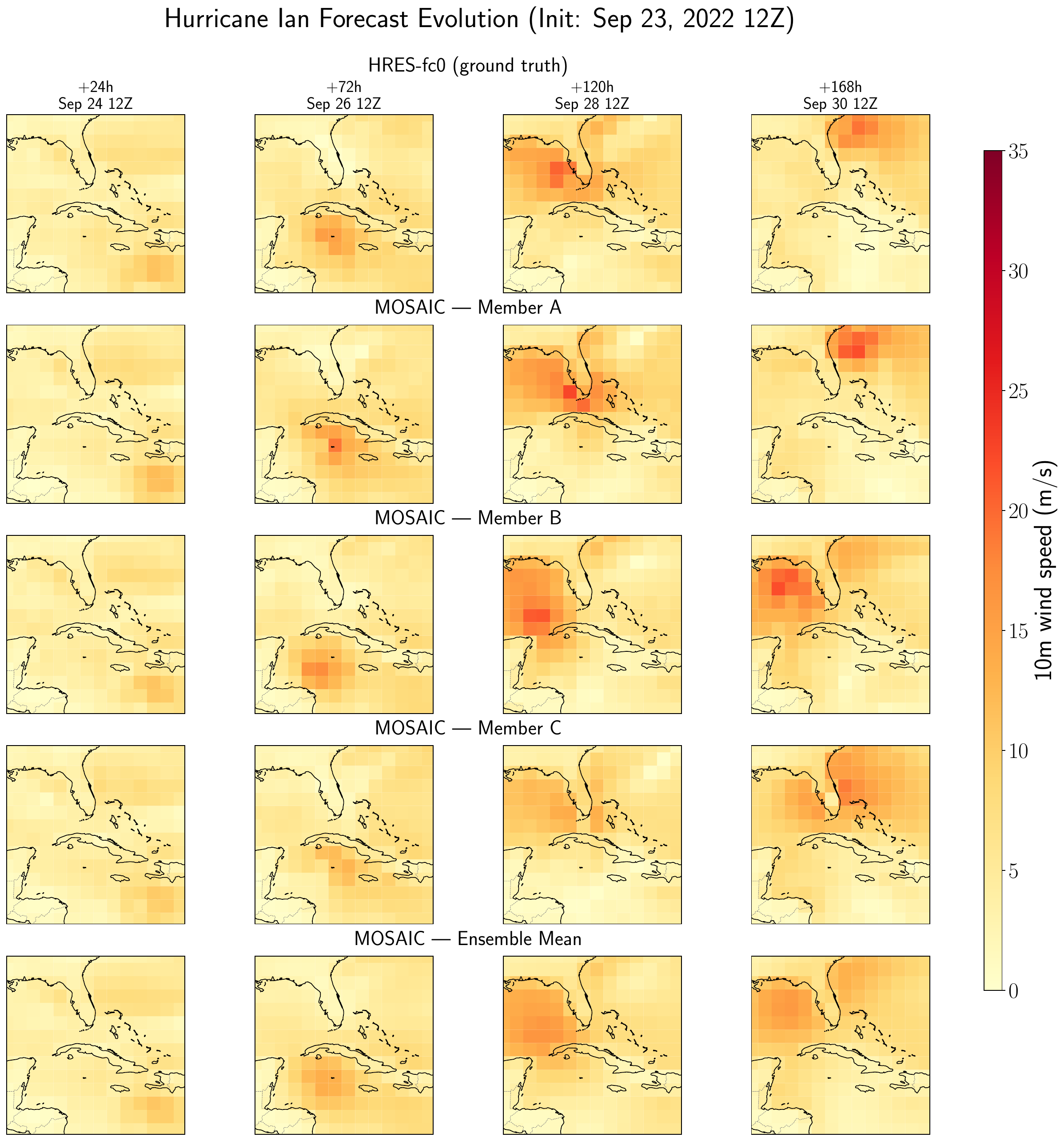}
    \caption{Hurricane Ian wind field evolution. Init: Sep~23, 2022 12Z. Columns show lead times (+24\,h, +72\,h, +120\,h, +168\,h). Rows: HRES-fc0 ground truth, three individual ensemble members, and ensemble mean (48 members). Variable: 10-meter wind speed (m/s). Region: Gulf of Mexico / Caribbean / southeastern US.}
    \label{fig:hurricane_ian_evolution}
\end{figure}

Fig.~\ref{fig:hurricane_ian_evolution} shows the 10-meter wind speed evolution from a single \textsc{Mosaic} run initialized on September~23, 2022, 12Z (5 days before Hurricane~Ian's Category~4 landfall). Ground truth (HRES-fc0 analysis), three individual ensemble members, and the ensemble mean are shown at four lead times (+24\,h, +72\,h, +120\,h, +168\,h). Individual members produce distinct realizations of the wind field, while the ensemble mean is smoother, particularly at longer lead times. The intensification and northward progression of the cyclone are visible in both the ground truth and the ensemble members, demonstrating that \textsc{Mosaic} captures the large-scale evolution of this extreme event despite operating at 1.5$^\circ$ resolution.

\subsection{Global Mass Conservation}
\label{app:mass_conservation}

\begin{figure}[htpb]
    \centering
    \includegraphics[width=0.7\columnwidth]{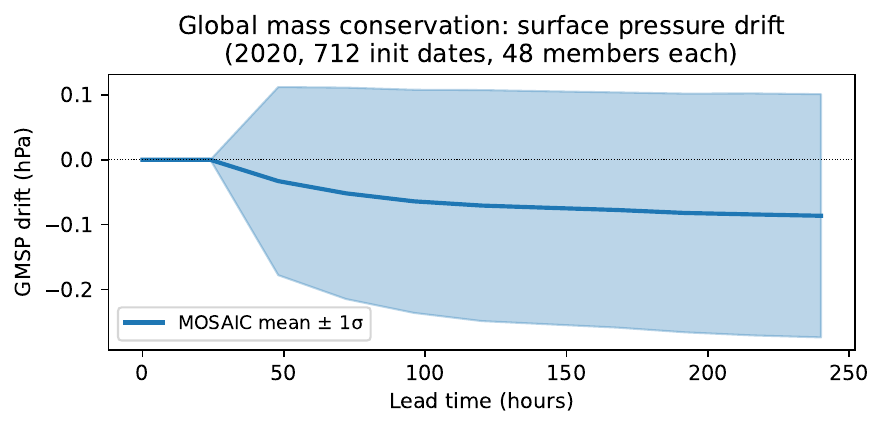}
    \caption{GMSP drift (mean $\pm$ 1$\sigma$) over lead time across 34,176 ten-day rollouts. The drift remains below 0.1\,hPa throughout, indicating stable mass conservation.}
    \label{fig:gmsp_drift}
\end{figure}

To assess model stability, we evaluate the global mean surface pressure (GMSP) drift over 10-day forecast rollouts. GMSP is computed as the area-weighted (cosine-latitude) global mean of mean sea level pressure. We use the ERA5-trained model with a 24\,h prediction step (the same model as the 1.5$^\circ$ benchmark in Section~\ref{section:experiments}) and evaluate over the 2020 test year: 712 initialization dates (00:00 and 12:00 UTC, year 2020) with 48 ensemble members each.
Table~\ref{tab:gmsp_drift} reports the GMSP drift relative to the initial condition. The maximum mean drift after 10 days is $-0.086$\,hPa ($0.009\%$ relative to ${\sim}1013$\,hPa), confirming that \textsc{Mosaic} neither systematically creates nor destroys atmospheric mass over extended rollouts. Fig.~\ref{fig:gmsp_drift} shows the drift trajectory.

\begin{table}[htpb]
\centering
\small
\begin{tabular}{lccc}
\toprule
\textbf{Lead time} & \textbf{Mean drift (hPa)} & \textbf{Relative} & \textbf{Max member} \\
\midrule
24\,h  & $-0.008 \pm 0.045$ & 0.001\% & $< 0.2$\,hPa \\
120\,h & $-0.042 \pm 0.112$ & 0.004\% & $< 0.5$\,hPa \\
240\,h & $-0.086 \pm 0.188$ & 0.009\% & $< 1.0$\,hPa \\
\bottomrule
\end{tabular}
\caption{GMSP drift over 10-day rollouts (2020 test year, 712 init dates $\times$ 48 members).}
\label{tab:gmsp_drift}
\end{table}

\newpage
\section{Implementation Details}
\label{app:implementation}

\subsection{Computational Resources}
\label{app:resources}

Table~\ref{tab:model_comparison} compares \textsc{Mosaic} with baseline MLWPs. All experiments were conducted on 8$\times$ NVIDIA H100 GPUs. Training requires 2 days (16 GPU-days total) using float16 precision throughout.

\begin{table}[htpb]
\centering
\small
\begin{tabular}{lccccc}
\toprule
\textbf{Model} & \textbf{Params} & \textbf{Precision} & \textbf{Resolution} & \textbf{Training} & \textbf{Duration} \\
\midrule
GraphCast & 36M & fp32 & 0.25° & 32 TPUv4 & 4 weeks \\
GenCast & 57.5M & fp32 & 0.25° & 32 TPUv5 & 5 days \\
FGN & 57.5M & fp32 & 0.25° & 490 TPUv5p/v6e & 3 days* \\
Pangu & 277M & fp32 & 0.25° & 192 V100 & 15 days \\
Aurora & 1,259M & bf16 & 0.25° & 32 A100 & 2.5 weeks \\
\midrule
\textsc{Mosaic} & 214M & fp16 & 1.5° & 8 H100 & 2 days \\
\bottomrule
\end{tabular}
\\[0.5em]
\raggedright
\caption{Comparison of \textsc{Mosaic} with baseline MLWPs. \textsc{Mosaic} achieves competitive performance despite operating at coarser resolution with significantly reduced computational resources.}
\label{tab:model_comparison}
\footnotesize *Total of 1,470 TPU-days compute (490 TPUs $\times$ 3 days)
\end{table}

\subsection{Data Description}
\label{app:data}

Table~\ref{tab:variables} lists all input and output variables used by \textsc{Mosaic}. The model takes as input 4 surface-level variables and 6 pressure-level variables at 13 pressure levels, yielding $4 + 6 \times 13 = 82$ dynamic channels per timestep. Additionally, 3 static fields are concatenated along the feature axis together with Cartesian coordinates $(x, y, z)$ on the unit sphere and 4 sinusoidal time embeddings (sine and cosine of day-of-year and year progress).

\begin{table}[htpb]
\centering
\small
\begin{tabular}{llll}
\toprule
\textbf{Variable} & \textbf{Type} & \textbf{Unit} & \textbf{Channels} \\
\midrule
\multicolumn{4}{l}{\textit{Surface-level variables (dynamic)}} \\
\midrule
2-meter temperature & Surface & K & 1 \\
10-meter U-wind & Surface & m/s & 1 \\
10-meter V-wind & Surface & m/s & 1 \\
Mean sea level pressure & Surface & Pa & 1 \\
\midrule
\multicolumn{4}{l}{\textit{Pressure-level variables (dynamic)}} \\
\midrule
Geopotential & Pressure & m$^2$/s$^2$ & 13 \\
Specific humidity & Pressure & kg/kg & 13 \\
Temperature & Pressure & K & 13 \\
U-component of wind & Pressure & m/s & 13 \\
V-component of wind & Pressure & m/s & 13 \\
Vertical velocity & Pressure & Pa/s & 13 \\
\midrule
\multicolumn{4}{l}{\textit{Static variables}} \\
\midrule
Surface geopotential & Static & m$^2$/s$^2$ & 1 \\
Land-sea mask & Static & 0--1 & 1 \\
Soil type & Static & categorical & 1 \\
\midrule
\multicolumn{4}{l}{\textbf{Total dynamic channels per timestep: 82}} \\
\bottomrule
\end{tabular}
\caption{Complete list of variables used by \textsc{Mosaic}. Pressure-level variables are provided at 13 levels: \{50, 100, 150, 200, 250, 300, 400, 500, 600, 700, 850, 925, 1000\}\,hPa.}
\label{tab:variables}
\end{table}

\paragraph{Data normalization.} All dynamic variables are standardized per channel (i.e., per variable and pressure level) using the mean and standard deviation computed over all spatial positions and all training timesteps via Welford's online algorithm. Each input state is normalized as $\hat{x} = (x - \mu) / \sigma$, where $\mu$ and $\sigma$ are the per-channel training statistics. The same normalization is applied to target states: $\hat{y} = (y - \mu) / \sigma$. The model predicts normalized next states directly, and predictions are denormalized at inference time via $x = \hat{x} \cdot \sigma + \mu$. Static variables are normalized independently using their own spatial mean and standard deviation and augmented with Cartesian coordinates $(x, y, z)$ on the unit sphere. During pretraining, normalization statistics are computed on the ERA5 training split (1979--2018). During finetuning, they are recomputed on the HRES-fc0 analysis data (2016--2021).

\paragraph{Data splits.} Pretraining uses ERA5 reanalysis from 1979--2018. We evaluate under two benchmarks (see Section~\ref{section:experiments}). For the \textbf{1.5$^\circ$ benchmark}, we finetune on ERA5 2007--2019 following \citet{archesweather} and test on 2020. For the \textbf{0.25$^\circ$ benchmark}, we finetune on HRES-fc0 analysis from 2016--2021 and test on 2022. All data is sampled at 6-hourly temporal resolution.

\paragraph{Data source and resolution.} All data is obtained from the WeatherBench~2 repository \citep{wb2} as Zarr archives on Google Cloud Storage. ERA5 reanalysis is provided as \texttt{1959-2023\_01\_10-6h-240x121\_equiangular\_with\_poles\_conservative.zarr} and HRES-fc0 analysis as \texttt{2016-2022-6h-240x121\_equiangular\_with\_poles\_conservative.zarr}. Both datasets are conservatively remapped from their native grids to a $240 \times 121$ equiangular latitude-longitude grid (1.5° resolution) at 6-hourly intervals. No additional regridding or interpolation is applied to the data prior to model input.

\subsection{Model Architecture}
\label{app:architecture}

Table~\ref{tab:model_architecture} lists the detailed architecture configuration. The U-Net comprises two coarse-graining stages and a bottleneck, with encoder and decoder layers at each stage.

\begin{table}[htpb]
\centering
\small
\begin{tabular}{lcccccc}
\toprule
\textbf{Stage} & \textbf{nside} & \textbf{Dim} & \textbf{Heads} & \textbf{Enc. Depth} & \textbf{Dec. Depth} & \textbf{MLP Ratio} \\
\midrule
Stage 1 & 64 & 768 & 12 & 4 & 2 & 4.0 \\
Stage 2 & 32 & 1024 & 16 & 4 & 2 & 4.0 \\
Bottleneck & 16 & 1280 & 20 & 2 &  & 4.0 \\
\midrule
\multicolumn{7}{l}{\textbf{Global parameters}} \\
\midrule
\multicolumn{7}{l}{Total parameters: 214M} \\
\multicolumn{7}{l}{GQA ratio: 4, QKV compression ratio: 1} \\
\multicolumn{7}{l}{RoPE: enabled ($\theta = 10000$), QK norm: disabled} \\
\multicolumn{7}{l}{RMSNorm elementwise affine: disabled} \\
\multicolumn{7}{l}{History steps $T$: 2 (1.5$^\circ$) / 4 (0.25$^\circ$), Noise dim: 32, $k$-neighbors: 24} \\
\midrule
\multicolumn{7}{l}{\textbf{Block-sparse attention parameters}} \\
\midrule
\multicolumn{7}{l}{Block attention size: 1024} \\
\multicolumn{7}{l}{Sparse block size: 128} \\
\multicolumn{7}{l}{Sparse block count: 24 (stage 1), 12 (stage 2), 4 (bottleneck)} \\
\bottomrule
\end{tabular}
\caption{Model architecture details.}
\label{tab:model_architecture}
\end{table}

\paragraph{Input embedding.}
Each spatial location on the latitude-longitude grid constitutes one token. For a history of $T$ consecutive timesteps ($T{=}2$ for the 1.5$^\circ$ benchmark, $T{=}4$ for the 0.25$^\circ$ benchmark), the $C_{\mathrm{dyn}}{=}82$ dynamic channels at each grid point are concatenated along the channel dimension. This vector is further concatenated with the static variables augmented with Cartesian unit-sphere coordinates ($C_{\mathrm{static}}{=}6$) and a single target-time embedding ($C_{\mathrm{time}}{=}4$: sine and cosine of day-of-year and year progress for the prediction target time~\citep{bodnarFoundation2025}), yielding a total input dimension of $T \times C_{\mathrm{dyn}} + C_{\mathrm{static}} + C_{\mathrm{time}}$ per token (174 for 1.5$^\circ$, 338 for 0.25$^\circ$). A preprocess MLP (Linear $\rightarrow$ RMSNorm $\rightarrow$ SiLU $\rightarrow$ Linear $\rightarrow$ RMSNorm) projects this to the hidden dimension of Stage~1, after which features are interpolated to the HEALPix mesh.

\paragraph{HEALPix mesh.}
The HEALPix grid is generated with the \texttt{healpy} library~\citep{zonca2019healpy} using \emph{nested} pixel ordering, which groups spatially nearby pixels into contiguous memory locations. The resolution parameter $N_{\mathrm{side}}$ determines the total number of pixels as $12 N_{\mathrm{side}}^2$: Stage~1 operates at $N_{\mathrm{side}} = 64$ (49{,}152 pixels), Stage~2 at $N_{\mathrm{side}} = 32$ (12{,}288 pixels), and the bottleneck at $N_{\mathrm{side}} = 16$ (3{,}072 pixels).

\paragraph{Downsampling and upsampling.}
Transitions between resolution levels follow the HEALPix quad-tree hierarchy with a downsampling factor of $f{=}4$. During downsampling (Section~\ref{section:our_method}), $W^{\downarrow}_x \in \mathbb{R}^{d_{\mathrm{out}} \times 4 d_{\mathrm{in}}}$ projects the stacked child features and $W^{\downarrow}_p \in \mathbb{R}^{d_{\mathrm{out}} \times 12}$ provides a position-aware bias from the $4 \times 3$ relative Cartesian coordinates, followed by RMSNorm. During upsampling, $W^{\uparrow}_x \in \mathbb{R}^{4 d_{\mathrm{out}} \times d_{\mathrm{in}}}$ expands each coarse pixel to four fine pixels and $W^{\uparrow}_p \in \mathbb{R}^{4 d_{\mathrm{out}} \times 12}$ again provides a position-aware bias. The encoder skip connection is added element-wise after the upsampling projection, followed by RMSNorm.

\paragraph{Output head.}
After the final decoder stage, features are normalized via RMSNorm, interpolated from HEALPix back to the latitude-longitude grid (Eq.~\ref{eq:interpolation} with source and target reversed), and passed through a postprocess MLP (RMSNorm $\rightarrow$ Linear $\rightarrow$ SiLU $\rightarrow$ Linear). The final linear layer maps to $C_{\mathrm{dyn}} = 82$ output channels with no activation function. The model predicts a normalized next state directly; denormalization recovers physical units at inference time.

\paragraph{Cross-attention interpolation.}
In the interpolation module (Eq.~\ref{eq:interpolation}), queries are derived from geometry with $W_q \in \mathbb{R}^{d \times 3}$ acting on $L_2$-normalized relative Cartesian positions, while keys and values are derived from RMSNorm-normalized source features via $W_k, W_v \in \mathbb{R}^{d \times d}$. An output projection $W_o \in \mathbb{R}^{d \times d}$ is applied after the attention-weighted sum. All projections are fully learned; neighbor indices and relative positions are fixed after initialization.

\paragraph{Noise injection details.}
The noise vector $\mathbf{z} \in \mathbb{R}^{32}$ is sampled once per forward pass from $\mathcal{N}(\mathbf{0}, \mathbf{I})$ and transformed by a learned layer $W_z \in \mathbb{R}^{32 \times 32}$. In each cSwiGLU block (Section~\ref{section:our_method}), a per-layer projection $W_n \in \mathbb{R}^{d_{\mathrm{ff}} \times 32}$ maps $\mathbf{z}$ to the feed-forward hidden dimension. Both $W_z$ and all $W_n$ are initialized with near-zero weights $\sim \mathcal{N}(0, 0.01)$, so that noise injection has negligible effect at the start of training and its influence is learned gradually. The same $\mathbf{z}$ is broadcast to every transformer block across the encoder, bottleneck, and decoder, acting as a global latent variable. Different ensemble members receive independently sampled noise vectors.

\subsection{Training Schedule}
\label{app:training}

Table~\ref{tab:training_schedule} shows the complete training schedule. Both pretraining and finetuning use a step size of 24\,h for the 1.5$^\circ$ benchmark and 6\,h for the 0.25$^\circ$ benchmark. Finetuning uses progressively longer autoregressive rollouts in both cases.

\begin{table}[htpb]
\centering
\small
\begin{tabular}{lccccc}
\toprule
\textbf{Stage} & \textbf{Steps} & \textbf{AR Length} & \textbf{Learning Rate} & \textbf{Weight Decay} & \textbf{Schedule} \\
\midrule
Pretrain & 250k & 1 & 1e-3 & 1e-2 & Cosine (1e-3 $\to$ 1e-6), no warmup \\
\midrule
Finetune 1 & 30k & 1 & 1e-4 & 1e-2 & Cosine decay by 1e-2 \\
Finetune 2 & 10k & 2 & 1e-5 & 1e-2 & Cosine decay by 1e-2 \\
Finetune 4 & 5k & 4 & 5e-6 & 1e-2 & Cosine decay by 1e-2 \\
Finetune 8 & 2.5k & 8 & 1e-6 & 1e-2 & Cosine decay by 1e-2 \\
Finetune 12 & 2.5k & 12 & 1e-6 & 1e-2 & Cosine decay by 1e-2 \\
\bottomrule
\end{tabular}
\caption{Training schedule details.}
\label{tab:training_schedule}
\end{table}

\paragraph{Optimizer.}
We use Muon \citep{muon} with momentum $\beta = 0.95$, Nesterov acceleration enabled, 5 Newton-Schulz orthogonalization steps, and no weight decay beyond what is specified per stage in Table~\ref{tab:training_schedule}. The base learning rate is $0.02$; per-stage values are listed in Table~\ref{tab:training_schedule} and follow the cosine schedule described therein.

\paragraph{Learning rate warmup.}
Pretraining uses no warmup. All finetuning stages employ a 500-step linear warmup from $10^{-6} \times \eta$ to the stage-specific learning rate $\eta$, followed by cosine annealing.

\paragraph{Early stopping.}
We apply early stopping based on validation loss at each stage. For pretraining, the criterion is the single-step prediction loss (24\,h for the 1.5$^\circ$ benchmark, 6\,h for the 0.25$^\circ$ benchmark). For finetuning stage $k$ (with $k$ autoregressive rollout steps), the criterion is the loss on the $k$-th predicted step. The best checkpoint from each stage initializes the next.

\paragraph{Gradient clipping.}
We clip gradient norms to a maximum of $1.0$ at every training step across all stages.

\paragraph{Distributed training.}
All experiments use distributed data parallelism (DDP) across 8 NVIDIA H100 GPUs with a per-GPU batch size of 2 (effective batch size 16). Training is performed entirely in float16 precision.

\paragraph{Weight initialization.}
All linear layers are initialized with $\mathcal{N}(0,\, \sigma)$ where $\sigma = \frac{1}{\sqrt{d_{\mathrm{in}}}} \min\!\big(1,\, \sqrt{d_{\mathrm{out}} / d_{\mathrm{in}}}\big)$, and all biases are set to zero. Residual-path layers -- the gate portion of SwiGLU ($W_{13b}$), the attention output projection ($W_o$), noise bias projections ($W_n$), noise generator ($W_z$), and upsampling projections -- are initialized with $\mathcal{N}(0,\, 0.01)$ so that residual contributions are near-identity at the start of training.

\paragraph{Autoregressive rollout during finetuning.}
Finetuning stages with $k > 1$ autoregressive steps use the pushforward trick~\citep{brandstetter2022message}: the first $k{-}1$ rollout steps are computed without gradients (i.e., with stopped gradients), and only the final $k$-th prediction step receives gradients. This avoids backpropagation through the full rollout chain, keeping memory requirements constant regardless of the number of rollout steps.

\paragraph{Training ensemble generation.}
During training, we use an ensemble of size $N{=}2$. At the first autoregressive step, the input state is replicated $N$ times within a single forward pass, and each replica receives an independently sampled noise vector $\mathbf{z}$, producing two ensemble members. For subsequent rollout steps, no further branching occurs: each of the two members evolves independently, receiving its own independently sampled noise vector at each step. The two-member ensemble structure is thus maintained throughout the full rollout. The CRPS loss (Section~\ref{app:loss}) is computed over the resulting 2-member ensemble at the final rollout step (the only step receiving gradients; see autoregressive rollout paragraph above). At inference, the ensemble size is increased to 48.

\subsection{Loss Function}
\label{app:loss}

The training objective is the latitude-weighted, variable-weighted fair CRPS (Eq.~\ref{eq:crps}):
\begin{equation}
    \mathcal{L} = \frac{1}{|D|} \sum_{d \in D} \frac{1}{H W} \sum_{h,w} \sum_{i=1}^{C} \alpha_i \, \omega_{h} \, \text{CRPS}(\hat{x}_{i,h,w,d}^{1:N},\, \hat{y}_{i,h,w,d}),
\end{equation}
where $d$ indexes the batch, $(h,w)$ indexes spatial grid points on the $H \times W$ latitude-longitude grid, $i$ indexes the $C{=}82$ output channels, $\alpha_i$ is the per-channel variable weight, and $\omega_h$ is the latitude weight. Both predictions $\hat{x}$ and targets $\hat{y}$ are in standardized (zero-mean, unit-variance) space.

\paragraph{Variable-level loss weights.}
Following \citet{lam2023learning}, pressure-level variables are weighted proportionally to their pressure level $p$ (in hPa), normalized by the mean pressure across all 13 levels ($\bar{p} = 463.46$\,hPa): $\alpha(p) = p / \bar{p}$. This assigns higher weight to lower-tropospheric levels, reflecting their greater meteorological importance for surface weather. Surface variables receive fixed weights. Table~\ref{tab:loss_weights} lists all weights.

\begin{table}[htpb]
\centering
\small
\begin{tabular}{lc}
\toprule
\textbf{Variable / Level} & $\boldsymbol{\alpha_i}$ \\
\midrule
\multicolumn{2}{l}{\textit{Surface variables}} \\
\midrule
2-meter temperature & 1.0 \\
10-meter U-wind & 0.1 \\
10-meter V-wind & 0.1 \\
Mean sea level pressure & 0.1 \\
\midrule
\multicolumn{2}{l}{\textit{Pressure levels (shared across all 6 pressure-level variables)}} \\
\midrule
1000\,hPa & 2.157 \\
925\,hPa & 1.996 \\
850\,hPa & 1.834 \\
700\,hPa & 1.510 \\
600\,hPa & 1.294 \\
500\,hPa & 1.079 \\
400\,hPa & 0.863 \\
300\,hPa & 0.647 \\
250\,hPa & 0.539 \\
200\,hPa & 0.431 \\
150\,hPa & 0.324 \\
100\,hPa & 0.216 \\
50\,hPa & 0.108 \\
\bottomrule
\end{tabular}
\caption{Variable-level loss weights $\alpha_i$. All six pressure-level variables share the same per-level weight.}
\label{tab:loss_weights}
\end{table}

\paragraph{Latitude weighting.}
To account for the convergence of meridians toward the poles, each grid row at latitude $\phi_h$ is weighted proportionally to the area it represents:
\begin{equation}
    \omega_h = \frac{\cos(\phi_h)}{\frac{1}{H} \sum_{h'=1}^{H} \cos(\phi_{h'})},
\end{equation}
so that the weights average to unity across latitudes.

\paragraph{Loss normalization.}
The loss is computed in standardized space: both model outputs and targets are normalized per channel using the training-set mean and standard deviation (Section~\ref{app:data}). The mean in the loss is taken over all dimensions (batch, latitude, longitude, and channels), with $\alpha_i$ and $\omega_h$ acting as importance multipliers. Training uses mixed-precision (float16) with gradient scaling via PyTorch's \texttt{GradScaler}. No additional loss normalization or scaling is applied.

\subsection{Block-sparse attention implementation}
\paragraph{Kernel Design} We implement block-sparse attention in Triton, using the FLA~\citep{yang2024fla} implementation as a foundation and following the memory-efficient approach of FlashAttention~\cite{flash_attn}. The forward pass loads query blocks into SRAM and streams selected key-value blocks through, computing attention without materializing the full attention matrix. The backward pass computes gradients for keys and values by iterating over all query blocks and only loading those into memory that are connected to a given key-value pair. The original NSA implementation batches query heads sharing the same key-value head, requiring sufficiently large batch sizes\footnote{At least 16 along each dimension in Triton.} for the dot product operations to be executed efficiently on Tensor Cores. In BSA, operating at the block level relaxes this constraint: batching occurs naturally across queries within a block, which at sizes ${\geq}128$ satisfies tile requirements and enables arbitrary GQA group size.

\begin{figure}[htpb]
    \centering
    \usetikzlibrary{decorations.pathmorphing}
\pgfdeclarelayer{foreground}
\pgfsetlayers{main,foreground}

\definecolor{bsa1}{HTML}{4a1412} %
\definecolor{bsa2}{HTML}{a6261d} 
\definecolor{bsa3}{HTML}{e04e3e} 
\definecolor{bsa4}{HTML}{fbb4a6} %

\definecolor{nsa1}{HTML}{08306b} %
\definecolor{nsa2}{HTML}{2171b5} 
\definecolor{nsa3}{HTML}{6baed6} 
\definecolor{nsa4}{HTML}{c6dbef} %

\begin{tikzpicture}
    \usetikzlibrary{patterns}
    \pgfplotsset{
        my axis style/.style={
            width=\linewidth,
            height=4.7cm, %
            ybar stacked,
            bar width=10pt,
            ymin=0, ymax=360,
            enlarge x limits=0.15,
            xtick={0,1,2,3,4,5,6}, %
            xticklabels={8k, 16k, 32k, 64k, 128k, 256k, 512k},
            xmin=1.2, xmax=5.8, 
            ylabel={Runtime (ms)},
            xlabel={Sequence length},
            axis x line*=bottom,
            axis y line*=left,
            ymajorgrids=true,
            grid style={dashed, gray!30},
            ylabel style={font=\normalsize},
            xlabel style={font=\normalsize},
            tick label style={font=\normalsize},
        }
    }

    \begin{axis}[
        my axis style,
        bar shift=-6pt, 
    ]
        \addplot[fill=bsa1, draw=none] coordinates {
            (0, 0.521) (1, 1.111) (2, 2.294) (3, 4.665) (4, 9.556) (5, 19.558) (6, 39.331)
        };
        \addplot[fill=bsa2, draw=none, postaction={pattern=crosshatch dots, pattern color=white!70}] coordinates {
            (0, 0.097) (1, 0.156) (2, 0.282) (3, 0.558) (4, 1.184) (5, 2.514) (6, 5.851)
        };
        \addplot[fill=bsa3, draw=none, postaction={pattern=dots, pattern color=white!70}] coordinates {
            (0, 0.008) (1, 0.008) (2, 0.008) (3, 0.011) (4, 0.021) (5, 0.052) (6, 0.150)
        };
        \addplot[fill=bsa4, draw=none, postaction={pattern=north east lines, pattern color=white!70}] coordinates {
            (0, 2.357) (1, 5.037) (2, 10.443) (3, 21.662) (4, 44.130) (5, 90.265) (6, 179.089)
        };
    \end{axis}
        
    \begin{axis}[
        my axis style,
        bar shift=6pt, 
        axis y line=none, axis x line=none, 
        ylabel={}, xlabel={},
    ]
        \addplot[fill=nsa1, draw=none] coordinates {
            (0, 0.527) (1, 1.109) (2, 2.249) (3, 4.584) (4, 9.292) (5, 18.711) (6, 38.364)
        };
        \addplot[fill=nsa2, draw=none, postaction={pattern=crosshatch dots, pattern color=white!70}] coordinates {
            (0, 1.424) (1, 3.086) (2, 6.283) (3, 12.689) (4, 70.794) (5, 10) (6, 10)
        };
        \addplot[fill=nsa3, draw=none, postaction={pattern=dots, pattern color=white!70}] coordinates {
            (0, 0.129) (1, 0.087) (2, 0.225) (3, 0.803) (4, 3.129) (5, 12.056) (6, 46.456)
        };
        \addplot[fill=nsa4, draw=none, postaction={pattern=north east lines, pattern color=white!70}] coordinates {
            (0, 5.487) (1, 13.886) (2, 27.844) (3, 68.144) (4, 151.358) (5, 249.23) (6, 195.18) 
        };
        \addplot[fill=white, draw=none] coordinates {
            (0, 0) (1, 0) (2, 0) (3, 0) (4, 0) (5, 20) (6, 20)
        };
        \addplot[fill=nsa4, draw=none, postaction={pattern=north east lines, pattern color=white!70}] coordinates {
            (0, 0) (1, 0) (2, 0) (3, 0) (4, 0) (5, 10) (6, 10)
        };
    \end{axis}
             
    \begin{axis}[
        my axis style,
        bar shift=6pt,
        axis y line=none, axis x line=none,
        ylabel={}, xlabel={},
    ]
        \addplot[draw=none, forget plot] coordinates {(5,295)};
        
        \draw[decorate, decoration={zigzag, segment length=4pt, amplitude=1pt}, thick, gray!90] 
             (axis cs: 4.95, 300) -- (axis cs: 5.2, 300);
        \draw[decorate, decoration={zigzag, segment length=4pt, amplitude=1pt}, thick, gray!90] 
             (axis cs: 5.95, 300) -- (axis cs: 6.2, 300);
    
        \node[font=\bfseries\scriptsize, above] at (axis cs: 5.08, 310) {701};
        \node[font=\bfseries\scriptsize, above] at (axis cs: 6.08, 310) {2120};
        
        \node[
            draw=gray!50, 
            fill=white, 
            fill opacity=0.9, 
            anchor=north west, 
            inner sep=2pt
        ] at (rel axis cs: 0.02, 0.99) {
            \scriptsize
            \begin{tabular}{@{}c@{\hspace{3pt}}l@{}} 
                \textcolor{bsa2}{\rule{8pt}{8pt}} & \small BSA (Ours) \\[2pt]
                \textcolor{nsa2}{\rule{8pt}{8pt}} & \small NSA~\citet{nsa} \\[-3pt]
                \multicolumn{2}{@{}c@{}}{\textcolor{gray!50}{\rule{80pt}{0.5pt}}} \\[1pt]
                \textcolor{bsa1}{\rule{6pt}{6pt}}\textcolor{nsa1}{\rule{6pt}{6pt}} & \footnotesize Local \hspace{7.5pt}
                \textcolor{bsa2}{\rule{6pt}{6pt}}\textcolor{nsa2}{\rule{6pt}{6pt}} \footnotesize Top-k \\[2pt]
                \textcolor{bsa3}{\rule{6pt}{6pt}}\textcolor{nsa3}{\rule{6pt}{6pt}} & \footnotesize Coarse \hspace{3pt}
                \textcolor{bsa4}{\rule{6pt}{6pt}}\textcolor{nsa4}{\rule{6pt}{6pt}} \footnotesize Selective \\[-1pt]
            \end{tabular}
        };
    
    \end{axis}
    
\end{tikzpicture}
    \caption{Runtime comparison of native sparse attention (NSA) vs block-sparse attention (BSA, ours) on NVIDIA A4500. NSA implementation from \citet{yang2024fla}. BSA achieves consistent speedups across all sequence lengths.}
    \label{fig:runtime_comp_nsa_bsa}
\end{figure}
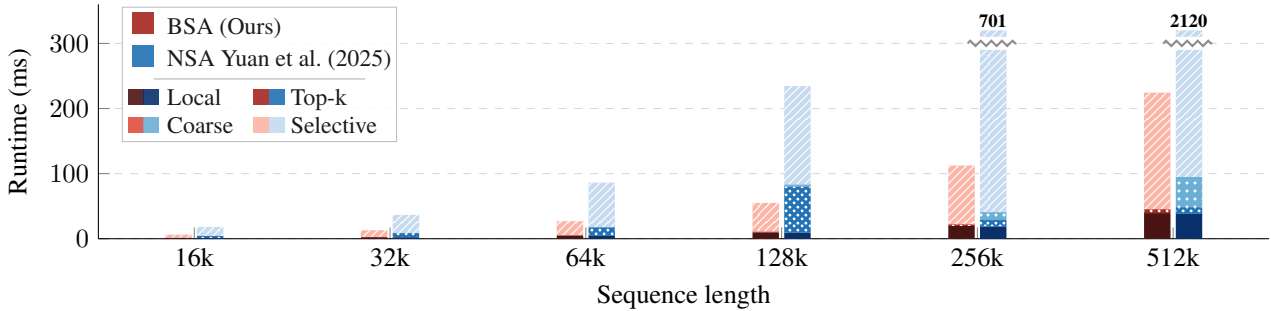

\paragraph{Block-sparse attention vs. native sparse attention.}
Figure~\ref{fig:runtime_comp_nsa_bsa} provides a per-component breakdown of BSA and NSA runtime. BSA's block-structured memory access patterns better utilize GPU tensor cores compared to NSA's irregular sparsity patterns, making BSA particularly suitable for high-resolution atmospheric grids. Table~\ref{tab:runtime_scaling} lists exact runtimes including dense attention (compiled \texttt{flex\_attention}); see Fig.~\ref{fig:runtime_scaling} for a visual comparison.

\begin{table}[htpb]
\centering
\small
\begin{tabular}{rrrrc}
\toprule
\textbf{Seq.\ len.} & \textbf{BSA} & \textbf{NSA} & \textbf{Dense} & \textbf{BSA/Dense} \\
\midrule
8k    &    2.97 &    7.52 &      2.68 & 0.9$\times$ \\
16k   &    6.31 &   17.95 &     11.20 & 1.8$\times$ \\
32k   &   13.13 &   36.49 &     44.75 & 3.4$\times$ \\
64k   &   27.84 &   86.04 &    176.85 & 6.4$\times$ \\
128k  &   57.15 &  234.12 &    752.74 & 13.2$\times$ \\
256k  &  114.72 &  700.32 &  3,383.69 & 29.5$\times$ \\
524k  &  230.68 & 2,115.22 & 14,258.72 & 61.8$\times$ \\
\bottomrule
\end{tabular}
\caption{Forward pass runtime (ms) for BSA, NSA, and dense attention across sequence lengths. Measured on NVIDIA RTX A4500 (batch=1, 16 heads, head\_dim=32). Dense uses compiled \texttt{flex\_attention}.}
\label{tab:runtime_scaling}
\end{table}

\subsection{Computational Efficiency}
\label{app:efficiency}

\paragraph{Training efficiency.}
As detailed in Table~\ref{tab:model_comparison}, \textsc{Mosaic} requires only 16 GPU-days total, operating entirely in float16 precision.
In contrast, GraphCast trains for 4 weeks on 32 TPUv4s, GenCast for 5 days on 32 TPUv5s, and FGN accumulates 490 TPU-days of compute, all using float32 precision.
The combination of lower resolution, the Muon optimizer, float16 precision, and block-sparse attention enables \textsc{Mosaic} to achieve competitive performance with substantially reduced compute relative to comparable probabilistic models.

\paragraph{Inference speed.}
Table~\ref{tab:inference_bench} compares inference cost on a single NVIDIA H100 GPU.
We report the amortized time per ensemble member per autoregressive step (\textbf{s/member/step}), measured from the largest ensemble that fits in memory, together with the peak GPU memory for that configuration.
\textsc{Mosaic} produces a 24-member, 10-step (240\,h) ensemble in 11.60\,s (0.048\,s/member/step), on par with the deterministic Stormer while generating full probabilistic forecasts.
Among other ensemble-capable models at comparable resolution, ArchesWeather-Mx4 is similarly fast but limited to 4 members, while ArchesWeatherGen requires 0.986\,s/member/step due to its diffusion-based sampling.

\begin{table}[htpb]
\centering
\small
\begin{tabular}{lcccrr}
\toprule
\textbf{Model} & \textbf{Res.} & \textbf{Params} & \textbf{Prec.} & \textbf{s/mem./step} & \textbf{Peak (GB)} \\
\midrule
\textsc{Mosaic} (Ours) & 1.5\textdegree & 214M & fp16 & 0.048 & 3.26 \\
ArchesWeatherGen & 1.5\textdegree & 384M & fp16 & 0.986 & 2.40 \\
ArchesWeather-Mx4 & 1.5\textdegree & 339M & fp16 & 0.058 & 2.08 \\
Stormer & 1.4\textdegree & 469M & fp16 & 0.062 & 11.24 \\
NeuralGCM$^\dagger$ & 1.4\textdegree & 12M & fp32 & 2.182 & 2.18 \\
\midrule
GraphCast$^\dagger$ & 0.25\textdegree & 36M & bf16 & 42.57 & 36.55 \\
GenCast$^{\dagger\S}$ & 0.25\textdegree & 57M & fp32 & 197.65 & 33.34 \\
Pangu-Weather$^\ddagger$ & 0.25\textdegree & -- & fp32 & 0.66 & 66.69 \\
\bottomrule
\end{tabular}
\\[0.5em]
\raggedright
\caption{Inference benchmarking on a single NVIDIA H100 GPU. \textbf{s/mem./step} is the wall-clock time per ensemble member per autoregressive step, computed from the largest ensemble run that fits in memory (24 members for \textsc{Mosaic}, ArchesWeatherGen, and NeuralGCM; 4 for ArchesWeather-Mx4; 1 for the rest). \texttt{torch.compile} is enabled for PyTorch models where applicable.}
\label{tab:inference_bench}
\footnotesize $^\dagger$\,JAX models. $^\S$\,GenCast uses diffusion-based sampling (1 noise level). \\
\footnotesize $^\ddagger$\,ONNX Runtime (no \texttt{torch.compile}).
\end{table}

\subsection{Evaluation Protocol}
\label{app:evaluation}

\paragraph{Inference ensemble generation.}
At inference, we generate a 48-member ensemble following the same branching scheme as during training (Section~\ref{app:training}): all members share the same initial condition and diverge at the first autoregressive step through independently sampled noise vectors. Each member then evolves autoregressively with a single noise sample per step.

\paragraph{Baseline results.}
All baseline scores reported in this paper are obtained from the WeatherBench~2 public evaluation framework \citep{wb2}. Before computing metrics, WeatherBench~2 first-order conservatively regrids all forecasts and ground truths to 1.5° resolution, ensuring a common evaluation grid. Since \textsc{Mosaic} operates natively at 1.5°, no additional regridding is required for our model. We do not re-run baseline models; we use their publicly available forecast outputs evaluated through WeatherBench~2.

\subsection{Reproducibility}
\label{app:reproducibility}

\paragraph{Software stack.}
All experiments are implemented in PyTorch~2.8 with CUDA~12.8. Custom block-sparse attention kernels are written in Triton~3.6. Key dependencies include \texttt{einops} (tensor rearrangement), \texttt{healpy}~\citep{zonca2019healpy} (HEALPix grid generation), \texttt{scikit-learn} (BallTree for neighbor lookup), \texttt{xarray} and \texttt{zarr} (data loading from WeatherBench~2), and \texttt{wandb} (experiment tracking). Spectral analysis uses \texttt{pyshtools} and \texttt{scipy}.

\paragraph{Code and data availability.}
Training and evaluation data are publicly available through the WeatherBench~2 repository \citep{wb2} on Google Cloud Storage.

\subsection{Aliasing in Compressive Weather Models}
\label{app:aliasing}

We elaborate on the second failure mode of spectral degradation: high-frequency artifacts arising from compressive encoding. The mechanism is aliasing, which has been studied in neural operators~\citep{stability_neural_ops, spectral_neural_ops} and in vision transformers~\citep{aliasfree_vit}. Here we summarize how it applies to weather models built on HEALPix and what we observe in our compression ablation.

\paragraph{Nyquist limit on the HEALPix grid.}
The HEALPix mesh at resolution $N_{\mathrm{side}}$ contains $12 N_{\mathrm{side}}^2$ pixels. \textsc{Mosaic} enters the U-Net at $N_{\mathrm{side}}{=}64$ (49{,}152~pixels), \textsc{Mosaic-C} at $N_{\mathrm{side}}{=}32$ (12{,}288~pixels). Halving $N_{\mathrm{side}}$ halves the number of representable spatial modes and therefore halves the Nyquist wavenumber of the latent grid. Features at finer scales than the latent Nyquist cannot be represented by the spatial layout of the coarse grid alone.

\paragraph{Aliasing mechanism.}
When the input is projected onto the coarse grid and then passed through pointwise nonlinearities, energy from frequencies above the coarse Nyquist limit folds back onto resolved frequencies~\citep{stability_neural_ops, spectral_neural_ops}. Once folded, this energy is indistinguishable from the resolved modes and is carried through the remaining nonlinear operations. Upon decoding back to the native grid, it re-emerges as spurious power at fine scales. The resulting spectral signature is an upward deviation of the model-to-reference ratio near the Nyquist limit -- the opposite of the damping produced by deterministic training.

\paragraph{Role of learnable interpolation.}
The Nyquist argument above describes a hard limit that applies when fine-scale information has nowhere to be stored other than across spatial pixels. \textsc{Mosaic} and \textsc{Mosaic-C} both project to the HEALPix mesh via the learnable cross-attention interpolation of Section~\ref{section:our_method}, which augments each coarse pixel with a multi-channel feature vector. Fine-scale variation can therefore be partially absorbed across the feature dimension rather than the spatial dimension, softening the hard Nyquist cut-off. As Fig.~\ref{fig:spectra_comparison}(a) shows, this offsets the onset of aliasing well beyond the formal $N_{\mathrm{side}}{=}32$ Nyquist wavenumber, but does not eliminate it: in \textsc{Mosaic-C}, the ratio still diverges upward in the highest resolved wavenumbers, where the channel capacity is no longer sufficient to encode the missing spatial detail.

\paragraph{Empirical signature.}
Fig.~\ref{fig:spectra_comparison}(a) reports kinetic energy spectral ratios at 24\,h, aggregated over the 2020 test year (720 initial conditions, 16 ensemble members for probabilistic models). \textsc{Mosaic} is stable across the spectrum, whereas \textsc{Mosaic-C} grows above 1.0 in the last wavenumbers before Nyquist. A qualitatively similar bump is visible for \textsc{GenCast} in the same figure, which also uses a coarse latent grid. Together with the 3.9\% nRMSE degradation of the spatial-compression ablation (Table~\ref{tab:ablation}\,g), this evidence indicates that compressive encoding induces a distinct architectural failure mode of spectral degradation, alongside the statistical damping discussed in the main text.

\begin{figure}[htbp]
    \centering
    \input{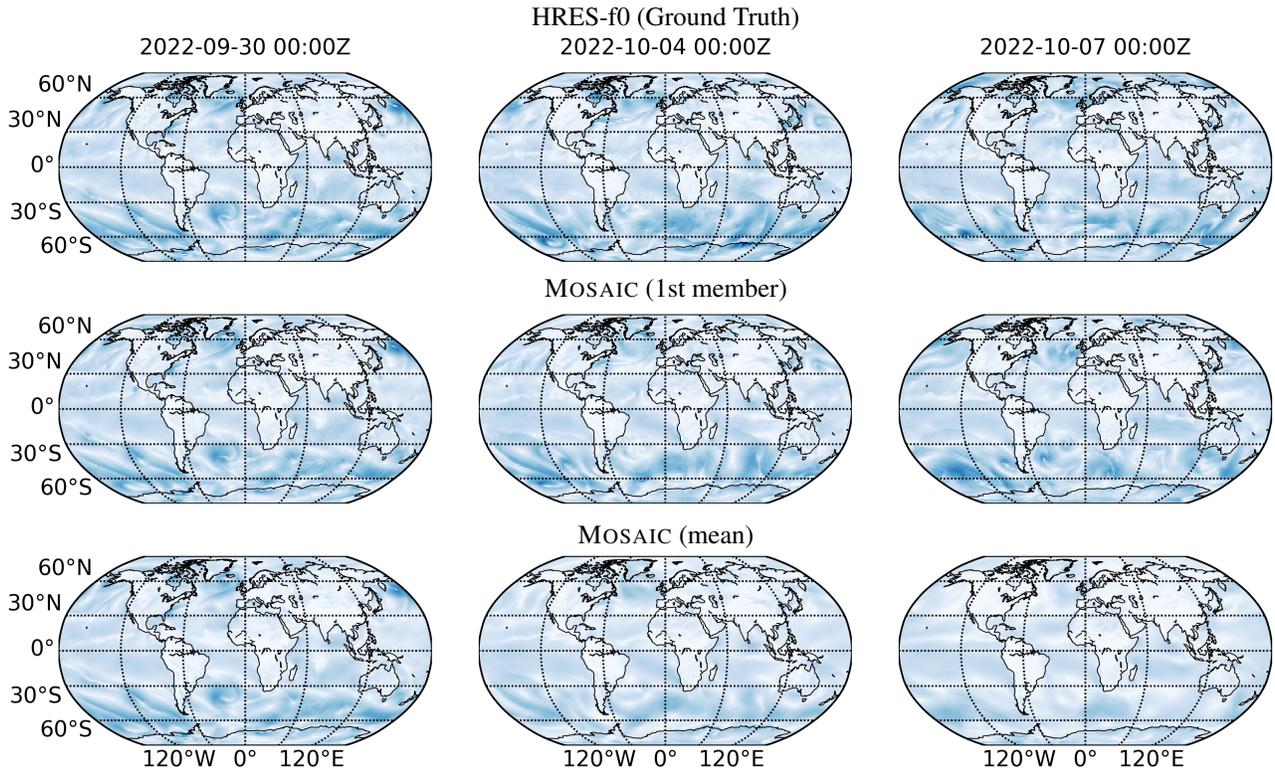}
    \caption{Forecast rollout trajectories showing 10-day evolution of wind speed fields at 850 hPa.}
    \label{fig:unrolling_wind}
\end{figure}

\begin{figure}[htbp]
    \centering
    \input{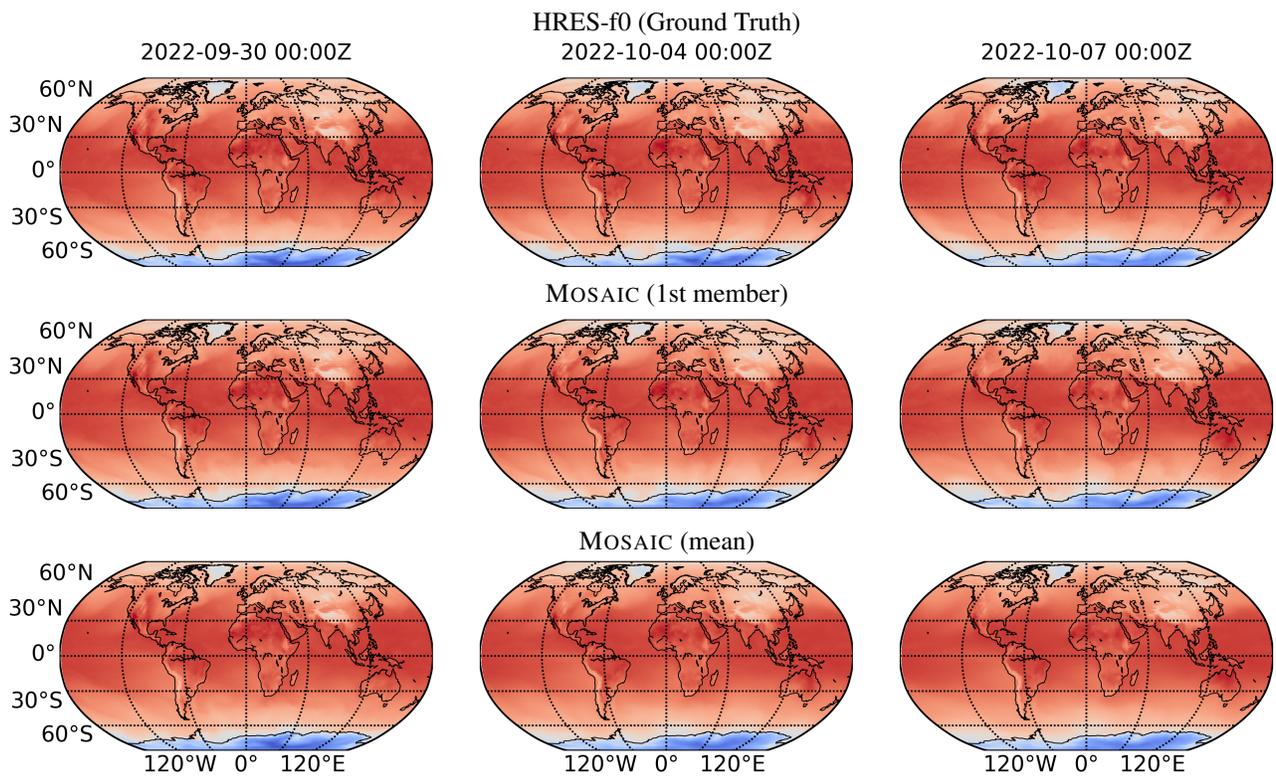}
    \caption{Forecast rollout trajectories showing 10-day evolution of surface temperature fields.}
    \label{fig:unrolling_temp}
\end{figure}

\begin{figure}[htpb]
    \centering
    \definecolor{myblue}{HTML}{4385BE}
\definecolor{myred}{HTML}{D14D41}
\definecolor{mylightred}{HTML}{F89A8A}
\definecolor{mygreen}{HTML}{879A39}
\definecolor{myyellow}{HTML}{DFB431}
\definecolor{myorange}{HTML}{DA702C}
\definecolor{myblack}{HTML}{1C1B1A}
\definecolor{mygray}{HTML}{575653}

\begin{tikzpicture}

    \pgfplotsset{
        spectrum_style/.style={
            width=0.36\linewidth,
            height=0.36\linewidth,
            xmode=log,
            ymode=log,
            xmin=30, xmax=125,
            ymin=3e-3, ymax=1e-1,
            grid=none,
            xtick=\empty,
            extra x ticks={40.03, 80.06, 120.15},
            extra x tick labels={1K, 500, 333},
            extra x tick style={
                tick align=outside,
                xtick pos=right,
                font=\tiny
            },
            every axis plot/.append style={line width=1.5pt},
            label style={font=\footnotesize},
            tick label style={font=\footnotesize},
            title style={font=\normalsize, yshift=-1ex}
        }
    }

    \begin{groupplot}[
        group style={
            group size=3 by 2,
            horizontal sep=0.5cm,
            vertical sep=1.0cm,
            y descriptions at=edge left,
            x descriptions at=edge bottom,
        },
        spectrum_style
    ]

    \nextgroupplot[
        title={IC: 2020-01-26},
        ylabel={Log (2m Temp)},
        legend to name=CommonLegendCombined,
        legend columns=-1,
        legend image post style={opacity=1},
        legend style={
            draw=none,
            fill=none,
            yshift=5pt,
            /tikz/every even column/.append style={column sep=0.5cm}
        }
    ]

        \addplot [dashed, color=mygray, line width=0.8pt, forget plot] table [x=n, y=ref_53] {figures/spectra/hres/ke/reference.dat};
        \node[anchor=south west, color=mygray, font=\tiny] at (axis cs: 90, 4e-2) {$n^{-5/3}$};

        \addplot [dashed, color=mygray, line width=0.8pt, forget plot] table [x=n, y=ref_3] {figures/spectra/hres/ke/reference.dat};
        \node[anchor=south west, color=mygray, font=\tiny] at (axis cs: 40, 5e-3) {$n^{-3}$};

        \addplot [color=mygray, opacity=0.8, solid, smooth, tension=0.7] table [x=n, y=spectrum] {figures/spectra/hres/ke/1/hres.dat};
        \addlegendentry{HRES-fc0}

        \addplot [color=myred, solid, smooth, tension=0.7] table [x=n, y=spectrum] {figures/spectra/hres/ke/1/mosaic_mean.dat};
        \addlegendentry{\textsc{Mosaic} (mean)}

        \addplot [color=mylightred, opacity=0.8, solid, smooth, tension=0.7] table [x=n, y=spectrum] {figures/spectra/hres/ke/1/mosaic_first.dat};
        \addlegendentry{\textsc{Mosaic} (1st member)}

    \nextgroupplot[
        title={IC: 2020-05-23},
    ]

        \addplot [dashed, color=mygray, line width=0.8pt] table [x=n, y=ref_53] {figures/spectra/hres/ke/reference.dat};
        \addplot [dashed, color=mygray, line width=0.8pt] table [x=n, y=ref_3] {figures/spectra/hres/ke/reference.dat};

        \addplot [color=mygray, opacity=0.8, solid, smooth, tension=0.7] table [x=n, y=spectrum] {figures/spectra/hres/ke/2/hres.dat};
        \addplot [color=myred, solid, smooth, tension=0.7] table [x=n, y=spectrum] {figures/spectra/hres/ke/2/mosaic_mean.dat};
        \addplot [color=mylightred, opacity=0.8, solid, smooth, tension=0.7] table [x=n, y=spectrum] {figures/spectra/hres/ke/2/mosaic_first.dat};

    \nextgroupplot[title={IC: 2020-07-14}]

        \addplot [dashed, color=mygray, line width=0.8pt] table [x=n, y=ref_53] {figures/spectra/hres/ke/reference.dat};
        \addplot [dashed, color=mygray, line width=0.8pt] table [x=n, y=ref_3] {figures/spectra/hres/ke/reference.dat};

        \addplot [color=mygray, opacity=0.8, solid, smooth, tension=0.7] table [x=n, y=spectrum] {figures/spectra/hres/ke/3/hres.dat};
        \addplot [color=myred, solid, smooth, tension=0.7] table [x=n, y=spectrum] {figures/spectra/hres/ke/3/mosaic_mean.dat};
        \addplot [color=mylightred, opacity=0.8, solid, smooth, tension=0.7] table [x=n, y=spectrum] {figures/spectra/hres/ke/3/mosaic_first.dat};

    \nextgroupplot[
        ylabel={Log (10m KE)},
    ]

        \addplot [dashed, color=mygray, line width=0.8pt] table [x=n, y=ref_53] {figures/spectra/hres/wind/1/reference.dat};
        \node[anchor=south west, color=mygray, font=\tiny] at (axis cs: 90, 4e-2) {$n^{-5/3}$};

        \addplot [dashed, color=mygray, line width=0.8pt] table [x=n, y=ref_3] {figures/spectra/hres/wind/1/reference.dat};
        \node[anchor=south west, color=mygray, font=\tiny] at (axis cs: 40, 5e-3) {$n^{-3}$};

        \addplot [color=mygray, opacity=0.8, solid, smooth, tension=0.7] table [x=n, y=spectrum] {figures/spectra/hres/wind/1/hres.dat};
        \addplot [color=myred, solid, smooth, tension=0.7] table [x=n, y=spectrum] {figures/spectra/hres/wind/1/mosaic_mean.dat};
        \addplot [color=mylightred, opacity=0.8, solid, smooth, tension=0.7] table [x=n, y=spectrum] {figures/spectra/hres/wind/1/mosaic_first.dat};

    \nextgroupplot[
        xlabel={Wavelength, km}
    ]

        \addplot [dashed, color=mygray, line width=0.8pt] table [x=n, y=ref_53] {figures/spectra/hres/wind/1/reference.dat};
        \addplot [dashed, color=mygray, line width=0.8pt] table [x=n, y=ref_3] {figures/spectra/hres/wind/1/reference.dat};

        \addplot [color=mygray, opacity=0.8, solid, smooth, tension=0.7] table [x=n, y=spectrum] {figures/spectra/hres/wind/2/hres.dat};
        \addplot [color=myred, solid, smooth, tension=0.7] table [x=n, y=spectrum] {figures/spectra/hres/wind/2/mosaic_mean.dat};
        \addplot [color=mylightred, opacity=0.8, solid, smooth, tension=0.7] table [x=n, y=spectrum] {figures/spectra/hres/wind/2/mosaic_first.dat};

    \nextgroupplot[]

        \addplot [dashed, color=mygray, line width=0.8pt] table [x=n, y=ref_53] {figures/spectra/hres/wind/1/reference.dat};
        \addplot [dashed, color=mygray, line width=0.8pt] table [x=n, y=ref_3] {figures/spectra/hres/wind/1/reference.dat};

        \addplot [color=mygray, opacity=0.8, solid, smooth, tension=0.7] table [x=n, y=spectrum] {figures/spectra/hres/wind/3/hres.dat};
        \addplot [color=myred, solid, smooth, tension=0.7] table [x=n, y=spectrum] {figures/spectra/hres/wind/3/mosaic_mean.dat};
        \addplot [color=mylightred, opacity=0.8, solid, smooth, tension=0.7] table [x=n, y=spectrum] {figures/spectra/hres/wind/3/mosaic_first.dat};

    \end{groupplot}

    \node[anchor=north] at ($(group c2r2.south) + (0,-1.0cm)$) {\ref{CommonLegendCombined}};

\end{tikzpicture}
    \caption{Examples of global temperature spectra at 2-meter height (top row) and kinetic energy spectra at 10-meter height (bottom row) for 10-day \textsc{Mosaic} forecasts compared to HRES-fc0 0.25$^\circ$ ground truth, shown for multiple initial conditions (all start at 00:00 UTC).}
    \label{fig:lsr_ex}
\end{figure}
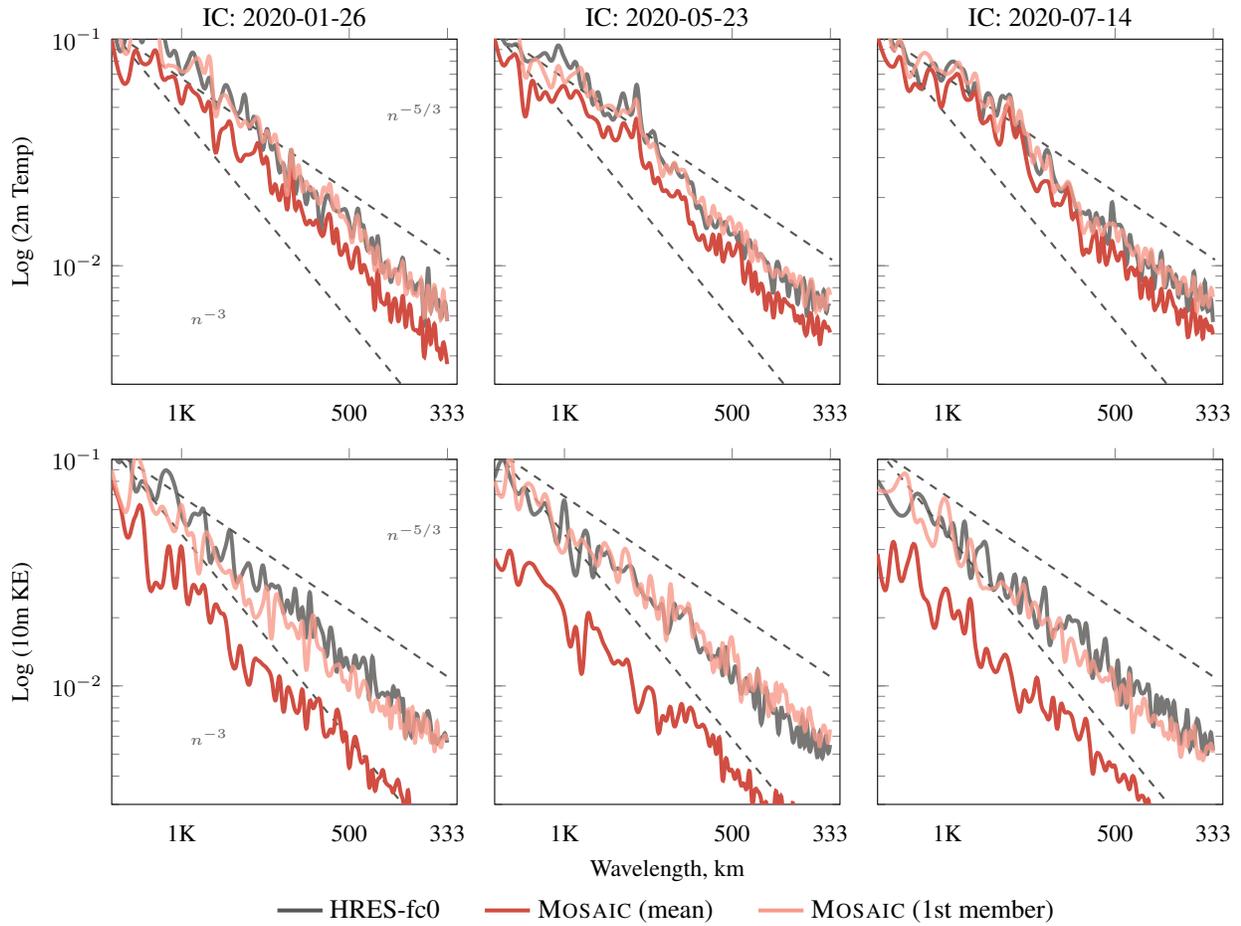

\end{document}